\documentclass{article}
\usepackage[utf8]{inputenc}
\usepackage[margin=1in]{geometry}
\usepackage{setspace}
\usepackage{amssymb}
\usepackage{amsmath}
\usepackage{amsthm}
\usepackage{bbm}
\usepackage{graphicx}
\usepackage{mwe}
\usepackage{color,soul}
\usepackage{multirow}

\usepackage{natbib}
\usepackage{authblk}
\usepackage{anyfontsize}
\usepackage{booktabs}

\usepackage{hyperref}
\hypersetup{
    colorlinks=true,
    linkcolor=blue,
    filecolor=magenta,      
    urlcolor=cyan,
    allcolors=blue,
}

\title{Regurgitative Training: The Value of Real Data in Training Large Language Models}
\author[1]{Jinghui Zhang}
\author[2]{Dandan Qiao}
\author[3]{Mochen Yang}
\author[1]{Qiang Wei}
\affil[1]{School of Economics and Management, Tsinghua University}
\affil[2]{School of Computing, National University of Singapore}
\affil[3]{Carlson School of Management, University of Minnesota}

\date{Draft Date: 7/3/2024}

\begin{document}
\onehalfspacing
\maketitle

\begin{abstract}
    What happens if we train a new Large Language Model (LLM) using data that are at least partially generated by other LLMs? The explosive success of LLMs, such as ChatGPT and LLAMA, means that a substantial amount of content online will be generated by LLMs rather than humans, which will inevitably enter the training datasets of next-generation LLMs. In this paper, we evaluate the implications of such ``regurgitative training" on LLM performance. Through fine-tuning GPT-3.5 with data generated either by itself or by other LLMs in a machine translation task, we find strong evidence that regurgitative training clearly handicaps the performance of LLMs. The ease of getting large quantities of LLM-generated data cannot compensate for performance loss –- even training with a fraction of real data is enough to outperform regurgitative training. The same performance loss of regurgitative training is observed on transformer models that we train from scratch. We carry out textual analyses to compare LLM-generated data with real human-generated data, and find suggestive evidence that the performance disadvantage of regurgitative training can be attributed to at least two mechanisms: (1) higher error rates and (2) lower lexical diversity in LLM-generated data as compared to real data. Based on these mechanisms, we propose and evaluate three different strategies to mitigate the performance loss of regurgitative training. In the first strategy, we devise data-driven metrics to gauge the quality of each LLM-generated data instance, and then carry out an ordered regurgitative training process where high-quality data are added before low-quality ones. In the second strategy, we combine data generated by multiple different LLMs (as an attempt to increase lexical diversity). In the third strategy, we train an AI detection classifier to differentiate between LLM- and human-generated data, and include LLM-generated data in the order of resemblance to human-generated data. All three strategies can improve the performance of regurgitative training to some extent but are not always able to fully close the gap from training with real data. Our results highlight the value of real, human-generated data in training LLMs, which cannot be easily substituted by synthetic, LLM-generated data. Given the inevitability of having some LLM-generated data in the training sets of future LLMs, our work serves as both a cautionary tale of its performance implication as well as a call-to-action for developing effective mitigation strategies.

\end{abstract}

\doublespacing

\section{Introduction} \label{sec:introduction}
Large language models (LLMs) are trained on inexplicably large amounts of data. Although the exact training datasets are undisclosed, popular LLMs such as ChatGPT, LLAMA, Claude, and Mistral are believed to have been trained on a combination of content on public Internet (e.g., Common Crawl), proprietary datasets licensed from third-parties, as well as crowd-generated data \citep{brown2020language,ouyang2022training,achiam2023gpt,touvron2023llama,anthropic2024claude}. With their explosive successes come widespread adoption -- people use LLMs in an ever increasing set of tasks, including writing \citep{noy2023experimental,chen2023large}, coding \citep{chen2021evaluating}, knowledge management \citep{lewis2020retrieval}, scientific discovery \citep{bran2023chemcrow,vert2023will}, and many more.

A natural consequence of such pervasive use is that, going forward, a substantial amount of content online will be created (at least partially) by LLMs. When building the next-generation LLMs, data generated by existing LLMs are likely to enter the training datasets. This produces a scenario which we refer to as \textit{Regurgitative Training}, where a new LLM is trained using data that are at least partially generated by itself or other LLMs. The overarching question we seek to answer in this paper is: \textbf{how does regurgitative training affect the performance of LLMs?}

Regurgitative training may be \textit{inevitable}. Indeed, there is evidence suggesting that a large part of the open web is already generated by machine translation models \citep[LLMs included,][]{thompson2024shocking}. Even data that are supposed to be human-generated (e.g., manual labels on crowdsourcing platforms) are often generated by LLMs \citep{veselovsky2023artificial}. As LLMs get better, it will be increasingly hard to distinguish between LLM-generated data from human-generated data post-hoc \citep{yang2023survey}. Additionally, some LLM developers have explicitly chosen to inject LLM-generated data into their training datasets, and empirically examine whether doing so can boost performance. For example, Apple acknowledges that its multi-modal LLM named MM1 has been trained on instruction-response pairs generated from GPT-4 and LLAMA\citep{mckinzie2024mm1}.

A priori, the impact of regurgitative training on LLM performance is unclear. On one hand, it represents an appealing opportunity to obtain large quantities of synthetic training data at relatively low costs, thereby offering a data quantity advantage. On the other hand, however, LLM-generated data may have lower quality than real, human-generated data -- they may contain more errors \citep{shumailov2023curse} or suffer more from the ``hallucination problem" \citep{rawte2023survey}.\footnote{Throughout the paper, we use ``real data" or "real human-generated data", in contrast with synthetic LLM-generated data, to refer to data that are generated by an organic process (typically by humans). Importantly, we do not assume real, human-generated data to be completely error-free; instead, we are interested in the \textit{comparison} of real vs. synthetic data in model training.} Overall, the performance impact of regurgitative training is jointly affected by both quantity and quality of synthetic data. In fact, although major players in the LLM arena, including Microsoft, Google and Meta, are all reported to use synthetic data for LLM training, the practice has garnered doubts from mainstream media.\footnote{Sources: \href{https://www.bloomberg.com/news/newsletters/2024-05-02/microsoft-google-and-meta-bet-on-fake-data-to-train-ai-models}{Microsoft, Google and Meta Bet on Fake Data to Build AI Models}, \href{https://www.wsj.com/tech/ai/the-ai-revolution-is-already-losing-steam-a93478b1}{The AI Revolution Is Already Losing Steam}, \href{https://www.wsj.com/tech/ai/ai-training-data-synthetic-openai-anthropic-9230f8d8}{For Data-Guzzling AI Companies, the Internet Is Too Small}, and \href{https://www.scientificamerican.com/article/ai-generated-data-can-poison-future-ai-models/}{AI-Generated Data Can Poison Future AI Models}.}

To understand the performance implications of regurgitative training, we carry out analyses under two different settings: \textit{fine-tuning} and \textit{training from scratch}, both of which represent realistic practice in building LLMs. First, using machine translation as an example generative task, we fine-tune the GPT-3.5 model with data generated by (i) GPT-3.5 itself, (ii) another LLM, namely GPT-4 or LLAMA2, and (iii) ground-truth real data. We then compare the out-of-sample translation performance between models fine-tuned with LLM-generated data versus those fine-tuned with real data. Second, we also build smaller-scale transformer-based models from scratch and repeat the above regurgitative training experiments. This is done for both machine translation and another generative task -- Q\&A -- to enhance the generalizability of our findings.

Across different generative tasks and model settings, we consistently observe that LLMs with regurgitative training \textit{underperform} those trained with real data. Given the same base model, training with more real data typically improves performance, whereas training with more LLM-generated data leads to quickly plateaued performance or even performance drops. In other words, the presumed performance advantage due to accessing large quantities of synthetic data is unrealized. Moreover, training with even a small proportion of real data is enough to outperform training only with LLM-generated data. The performance disadvantages of regurgitative training are especially pronounced when the LLM responsible for generating training data is not good at the task. 

We also perform several textual analyses to make sense of regurgitative training's performance disadvantages. As can be expected, errors in LLM-generated data is one of the culprits. Interestingly, we find evidence that it may not be the only contributing mechanism. Specifically, LLM-generated data exhibit lower degrees of lexical diversity than real data, echoing several recent research \citep{padmakumar2023does,doshi2023generative,anderson2024homogenization} in the contexts of academic writing or creative ideation. Lower lexical diversity in LLM-generated data also partially explains the performance disadvantages of regurgitative training.

In light of these findings, we propose and evaluate three different strategies to more carefully leverage LLM-generated data in regurgitative training. The first strategy borrows from the semi-supervised learning literature and prioritizes using high-quality LLM-generated data over low-quality data, where ``quality" is gauged either by prediction confidence or by an external supervised learning model. Second, as an attempt to address the diversity deficit of LLM-generated data, we combine data generated by a mixture of different LLMs in training. Finally, the third strategy makes use of ``AI detectors", i.e., classification models that try to distinguish between LLM- vs. human-generated content. We train and deploy a capable AI detector model on LLM-generated data, and use LLM-generated data in regurgitative training in the order of predicted probability of being generated by humans (i.e., prioritizing LLM-generated data that resemble human-generated data). Our results demonstrate that all three strategies have some power to improve the performance of regurgitative training. In a few cases with transformer models trained from scratch, the performance improvements are quite significant. Meanwhile, effectiveness of these strategies tends to be small in the fine-tuning setting, and none of them can fully close the gap from training with real data. This further highlights the unique value of real, human-generated data in LLM training.

Our work makes several contributions to the fast growing literature on generative AI and LLMs. Aside from all the amazing capabilities of modern LLMs, we offer a sobering analysis of regurgitative training, which may become inevitable as LLMs get more deeply integrated into various content generation tools and channels. Our empirical evidence demonstrates that regurgitative training stalls or hurts LLM performance, because LLM-generated data, as coherent or convincing as they may seem, still fall short of real data. Therefore, more productive regurgitative training requires a more careful use of LLM-generated data. The three mitigation strategies we propose and test represent practical design artifacts that can mitigate the performance loss associated with regurgitative training. In the meantime, the fact that no mitigation strategy we have explored can catch up with the performance of training with real data is an indication that real data remain one of the most valuable assets of LLM training, and cannot be easily substituted by synthetic data produced by existing LLMs.

\section{Relevant Literature} \label{sec:literature}
Our work is closely related to self training in the semi-supervised learning literature and data augmentation in the deep learning literature, both of which are briefly reviewed in this section. As will be discussed later, although regurgitative training in LLMs is fundamentally different from the conventional schemes of self training or data augmentation, both offer some valuable ideas that can inform our understanding of regurgitative training as well as potential approaches to manage its performance downsides.

\subsection{Self Training} \label{sec:literature_self_training}
Self training is one of the classic approaches in semi-supervised learning to train a machine learning model using both labeled and unlabeled data \citep{scudder1965probability,nigam2000analyzing}. Take classification tasks as an example, the idea is to first build a model on the labeled data via standard supervised learning procedures, obtain the model's predictions on the unlabeled data, then take the most confident predictions (e.g., data instances with most extreme predicted probabilities) and treat them as additional labeled data to re-train the model. As a way to convert some unlabeled data into labeled data, self training is useful especially when the original labeled data are scarce. 

There is an extensive literature on self training, both in traditional machine learning \citep[see][for surveys of the topic]{pise2008survey,triguero2015self} and in modern deep learning \citep[e.g.,][]{xie2020self}. One key insight from this body of work is that the effectiveness of self training depends heavily on the ability to accurately estimate ``prediction confidence". Because predictions with high confidence are used as ``pseudo-labels", having accurate confidence scores imply that the ``pseudo-labels" are more likely to be correct (i.e., the same as ground-truth labels). 

Regurgitative training of LLMs resembles self training in that model-predicted pseudo labels are used to further train the model. However, it is unclear whether the conventional wisdom of self training would still apply in the case of regurgitative training, because of the difficulty in assessing confidence of LLM outputs \citep{lin2023generating}. In classification tasks, predicted class probabilities naturally serve as the measure to quantify the uncertainty in a classifier's predictions. However, LLMs are much more complex than classifiers -- they generate multi-token answers in response to prompts. Unless in highly restricted scenarios (e.g., evaluating a single-digit response to the question ``what is 2+2"), it is generally not straightforward to define or measure confidence in LLM outputs. As a result, current LLMs do not automatically produce confidence scores for their responses and uncertainty quantification in LLMs remain an open question with many ongoing research, including asking LLMs themselves for ``self-evaluation/reflection'' \citep{chen2023quantifying}, re-running the same prompt multiple times and measuring internal consistency among responses \citep{kotelanski2023methods}, and tapping into human expertise \citep{shankar2024validates}.

\subsection{Data Augmentation} \label{sec:literature_data_augmentation}
Data augmentation represents another approach to enrich potentially limited labeled data with synthetic data. It works by injecting noises into existing labeled data instances to artificially create new data instances that can be assigned the same labels. In language tasks, a common data augmentation strategy is back-translation \citep{yu2018qanet}, where a sentence is first translated into a different language and then reverted back to the original language to achieve paraphrasing. In vision tasks, data augmentation may involve image transformation techniques such as rotation, color / contrast modification, etc. \citep{cubuk2020randaugment,xie2020unsupervised}. These augmentation strategies can benefit model performance if the injected noises do not change the labels, thereby creating more training data with valid labels.

Although data augmentation is procedurally quite different from regurgitative training, it does offer an insight that can help enhance the performance of regurgitative training. \cite{xie2020unsupervised} found that data augmentation is more effective when the augmentation strategy can generate a diverse set of instances rather than only introducing small, local perturbations. Learning from a diverse set of augmented data can enable the model to achieve competitive performance with fewer examples. Conceptually, this finding is also consistent with observations made in other machine learning research outside of data augmentation \citep[e.g.,][]{gong2019diversity}, where the diversity of training data instances is positively associated with predictive performance. Later, we leverage this insight in one of the strategies designed to mitigate performance loss of regurgitative training, by mixing data generated by different LLMs as an attempt to introduce greater diversity to the training process.

\subsection{Regurgitative Training} \label{sec:literature_regurgitative}
Regurgitative training of generative AI models represents a new problem that has only begun to receive scholarly attention very recently. The earliest work we could identify is \cite{shumailov2023curse}, which documents that using model-generated data to train next-generation models can create irreversible performance losses, a phenomenon they term ``model collapse". They demonstrate this in common generative AI architectures such as variational autoencoders, Gaussian mixture models, and small-scale LLMs. Moreover, they provide theoretical intuitions that model collapse arises due to errors in model-generated data, which accumulates over more iterations of regurgitative training. Subsequently, the phenomenon of model collapse has also been observed in generative image models \citep{alemohammad2023self,bertrand2023stability}. 

In the meantime, efforts to mitigate model collapse are underway. \cite{bertrand2023stability} show that model collapse can be avoided if (i) the proportion of real data is sufficiently high and (ii) model-generated data approximate the distributions of real data well enough. Furthermore, \cite{gerstgrasser2024model} propose to alleviate model collapse by ``accumulating data"; that is, using the totality of real and model-generated data (rather than just the model-generated data) to train new models. 

We build upon this nascent stream of research and aim to make several distinct contributions. First, we consider regurgitative training of a LLM not only by data generated by itself, but also by other LLMs with varying degrees of capabilities. This is already taking place in practice \citep[e.g.,][]{mckinzie2024mm1} but has not been systematically explored in the literature. Second, prior work such as \cite{shumailov2023curse} focused on early versions of generative models (e.g., non-transformer-based models or small pre-trained models). Instead, we carry out comprehensive experiments with leading LLMs at the time of research (e.g., GPT-4 and LLAMA2) as well as transformer models trained from scratch, thereby providing a more up-to-date understanding of regurgitative training. Third, we conduct textual analysis to explore the mechanisms that may explain the performance disadvantages of regurgitative training. Finally, we propose several new mitigation strategies beyond what has been tested so far, and empirically evaluate their effectiveness.

\section{Performance Impact of Regurgitative Training} \label{sec:experiments}
In this section, we aim to understand how regurgitative training affects the performance of an LLM through two sets of experiments, respectively constructed to reflect two representative practices in LLM training: (i) fine-tuning and (ii) training from scratch. Fine-tuning allows users to adapt an existing LLM to their own use cases and, as mentioned before, is a widely adopted practice in the industry \citep[e.g.,][]{mckinzie2024mm1}. We expect a lot of LLM training will take the form of fine-tuning, because training a state-of-the-art LLM from scratch is highly complex and resource-intensive. Meanwhile, we also consider the case of training smaller-scale transformer language models from scratch, which may be necessary for companies that cannot leverage third-party LLMs due to data security and privacy issues. 

For both fine-tuning and training from scratch, we focus on machine translation as the generative task of interest. Translation represents a common application for LLMs, and the performance of a translation model can be evaluated with well-established standards and metrics in the literature. This enables us to robustly assess the performance variations resulting from regurgitative training. In the case of training from scratch, we also replicate the main findings with a different generative task, namely Q\&A.

\subsection{Experiments with Fine-Tuning} \label{sec:experiments_finetune}
To carry out LLM fine-tuning for translation, we use the Europarl parallel corpus \citep{koehn-2005-europarl}. Sourced from the proceedings of the European Parliament, the corpus contains parallel sentences in multiple European languages. We specifically use pairs of German-English sentences. After basic pre-processing steps (e.g., removing special HTML tags, eliminating noisy characters, and handling null values), we end up with 1,908,849 sentence pairs for our analyses. We treat these sentence pairs as \textit{real data}.

A popular and widely used metric to evaluate the performance of a translation model is the \textbf{BLEU} (BiLingual Evaluation Understudy) score \citep{papineni2002bleu}. It evaluates the quality of a model-generated translation (also called a ``hypothesis translation") in comparison to one or more reference translations.\footnote{The BLEU score can be analogously defined for a corpus of hypothesis translations and the corresponding corpus of reference translations. We describe the simpler case with a single hypothesis translation here for ease of understanding, and refer readers to \cite{papineni2002bleu} for the more general case.} The BLEU score is calculated as the n-gram overlap between the hypothesis translation and reference translations. It ranges from 0 to 1 and a higher BLEU score generally indicates better quality translations. Formally, the BLEU score is defined as
\begin{equation}
\label{eq:bleu}
\text{BLEU}  =\min \left\{1, \exp \left(1-\frac{r}{c} \right) \right\} \cdot 
 \exp \left(\sum_{n=1}^{N}w_{n}\log{p_{n}} \right)
\end{equation}
In the first term, $c$ is the length of hypothesis translation and $r$ is the ``effective" length of reference translations (defined as the length of the reference translation that best matches the hypothesis translation). This term serves as a ``brevity penalty" that assigns a higher score for a better match in lengths between hypothesis and reference translations. In the second term, $p_n$ denotes the $n$-gram precision and is defined as
\begin{equation}
\label{eq:pn}
p_{n}  =\frac{\sum_{n\mbox{-}gram \in hypothesis}Count_{matched}(n\mbox{-}gram)}{\sum_{n\mbox{-}gram \in hypothesis}Count(n\mbox{-}gram)} 
\end{equation}
where $Count(n\mbox{-}gram)$ counts the number of $n$-gram in the corpus and $Count_{matched}(n\mbox{-}gram)$ counts the number of $n$-gram matches between the hypothesis translation and reference translations. In Equation \eqref{eq:bleu}, the $n$-gram precision scores are then weighted by $w_n$ (e.g., uniform weighting $w_n = \frac{1}{N}$) to compute the overall BLEU score.

To implement and evaluate regurgitative fine-tuning, several components need to be defined first, including a baseline LLM to be fine-tuned, a set of training data (either real or generated by other LLMs) used for fine-tuning, and a fine-tuned LLM for evaluation. In our context, we use the GPT-3.5 model as the baseline LLM,\footnote{At the time of our research, OpenAI's fine-tuning service was restricted to the GPT-3.5 model.} then fine-tune it with (i) real human-generated data, (ii) data generated by GPT-3.5 itself, and (iii) data generated by two other LLMs, namely GPT-4 and LLAMA2. This creates four fine-tuned LLMs, all of which are evaluated on the same testing data for performance comparison.

More specifically, we randomly select 5,000 sentence pairs from the original corpus for fine-tuning and 10,000 sentence pairs as the testing data. When fine-tuning with real data, the 5,000 German sentences are used as inputs and the corresponding 5,000 English sentences are used as target translations. When fine-tuning with LLM-generated data, the same 5,000 German sentences are used as inputs, but the target translations are generated by the corresponding LLM. For GPT-3.5, GPT-4, and LLAMA2, we obtain their translations with the same system prompt: ``You are a chatbot that can translate German to English.", and the German sentences are given to the LLMs as user inputs. Using each set of data, we carry out progressive fine-tuning over five batches, adding 1,000 data instances per batch and recording translation performance on the testing data after each batch.

\begin{figure}[!tbh]
 \centering
 {\includegraphics[width=0.6\textwidth]{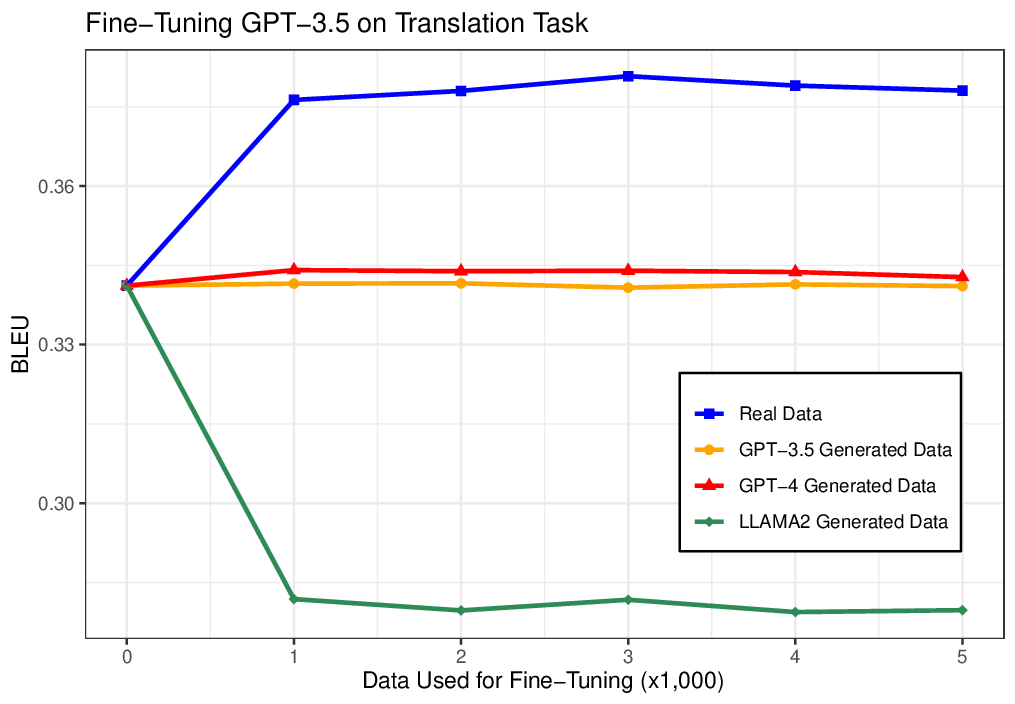}}
 \caption{Performance of Fine-Tuning GPT-3.5 Model \label{fig:gpt_performance}}
\end{figure}

We show the results in Figure \ref{fig:gpt_performance}. Each line represents the translation performance of a particular model over five fine-tuned batches. The $X$-axis indicates batch index, marking the number of data instances utilized in the fine-tuning process. The $Y$-axis represents the BLEU score, where a higher value corresponds to better translation performance.

From the figure, it is evident that the performance of fine-tuning with LLM-generated data (both from the baseline LLM itself and from other LLMs) clearly lags behind the performance of fine-tuning with real human-generated data. Moreover, regurgitative training with different LLMs have differential performance impact. Fine-tuning with data generated by GPT-3.5 itself does not significantly change performance, and fine-tuning with GPT-4 generated data only results in slight performance improvement over the baseline model (i.e., at point 0 on the $X$-axis). However, fine-tuning with LLAMA2 generated data significantly \textit{degrades} performance compared to the baseline. This is likely because the three LLMs have different translation capabilities. Since we have the ground-truth translations for the 5,000 fine-tuning data, we can directly compute the BLEU scores of translations generated by the three LLMs, and indeed find GPT-4 to be the best ($BLEU = 0.3454$), followed by GPT-3.5 ($BLEU = 0.3428$) and LLAMA2 ($BLEU = 0.2417$). 

These results underscore the overall underwhelming, and potentially detrimental effects of regurgitative training. Compared to training with real data, regurgitative training largely \textit{stalls learning}. Regurgitative training with a better-performing LLM improves performance only marginally and is not sufficient to catch up with the performance on real data. Worse yet, regurgitative training with a less capable LLM can significantly hurt performance. 

\begin{figure}[!tbh]
 \centering
 {\includegraphics[width=0.55\textwidth]{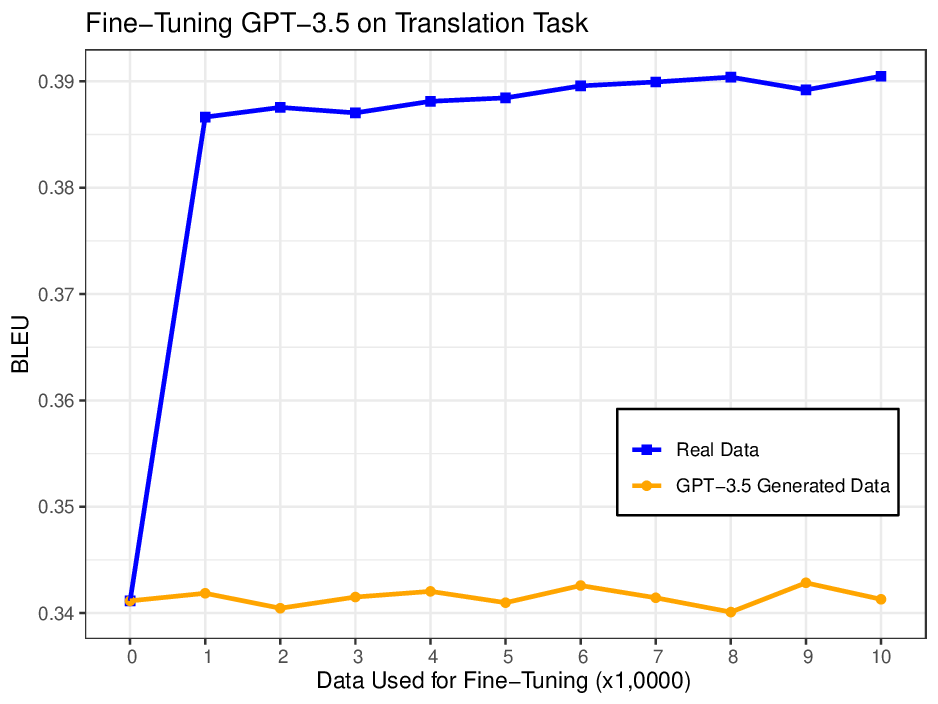}}
 \caption{Performance of Fine-Tuning GPT-3.5 Model (Augmented Data Size) \label{fig:gpt_performance1}}
\end{figure}

Note that in the above experiments, we use a small set of data for fine-tuning. This decision stems from the remarkable few-shot learning capabilities of modern LLMs \citep{brown2020language}. In addition, we also conduct a robustness check to understand whether the performance of regurgitative training may be different if more fine-tuning data are available. Specifically, we augment the fine-tuning data size by 20 times, to a total of 100,000 data instances, and incrementally add 10,000 per batch. For efficiency and cost considerations, we only run this size-augmented fine-tuning with real data and data generated by GPT-3.5 itself. We then evaluate each fine-tuned models on the same testing data as before. The results are presented in Figure \ref{fig:gpt_performance1}. We again observe that regurgitative training is unable to improve translation performance and substantially underperforms training with real data.

\subsection{Experiments with Models Trained from Scratch} \label{sec:experiments_self_train}
We now turn to training models from scratch and understanding the performance impact of regurgitative training in this case. Specifically, we build transformer models using the translation data. The transformer architecture serves as a foundational component powering the majority of modern LLMs, and has found extensive applications in machine translation and a variety of other natural language tasks \citep{vaswani2017attention,wolf2020transformers}. We therefore choose to train small-scale transformer models from scratch, as an attempt to approximate the practice of training transformer-based models without leveraging third-party LLMs. 

We follow \cite{vaswani2017attention} to build the baseline transformer models. Transformer has an encoder-decoder architecture, which uses stacked layers of multi-head self-attention and point-wise, fully connected feed-forward networks for both the encoder and decoder. It also employs a residual connection on each sub-layer, followed by layer normalization. For implementational details of these transformer elements, we refer to \cite{vaswani2017attention}. 

\begin{figure}[!tbh]
 \centering
 {\includegraphics[width=0.6\textwidth]{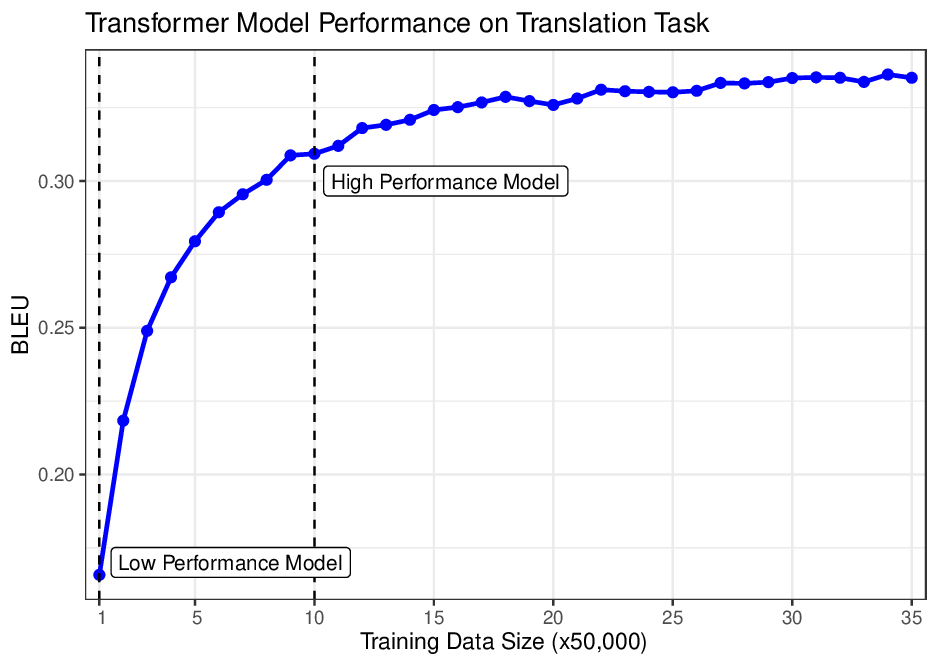}}
 \caption{Performance of Transformer Models with Varying Training Data Sizes \label{fig:translation_real}}
\end{figure}

As our previous fine-tuning results have shown, the performance impact of regurgitative training can vary with the capability of the model used to generate training data. Therefore, we train both a ``low-performance" and a "high-performance" baseline models. This is done by gradually adding 50,000 sentence pairs (randomly sampled from the German-English corpus) per batch for training, and evaluate the model's translation performance on a fixed testing dataset of 50,000 sentence pairs. As shown in Figure \ref{fig:translation_real}, we observe that the model's performance improves quickly with the initial increase in training data size, and saturates after being trained with sufficient data. We choose the model trained with 50,000 data instances (i.e., 1 batch) as our low-performance baseline and the one trained with 500,000 data instances (i.e., 10 batches) as the high-performance baseline. 

\begin{figure}[!tbh]
 \centering
  {\includegraphics[width=\textwidth]{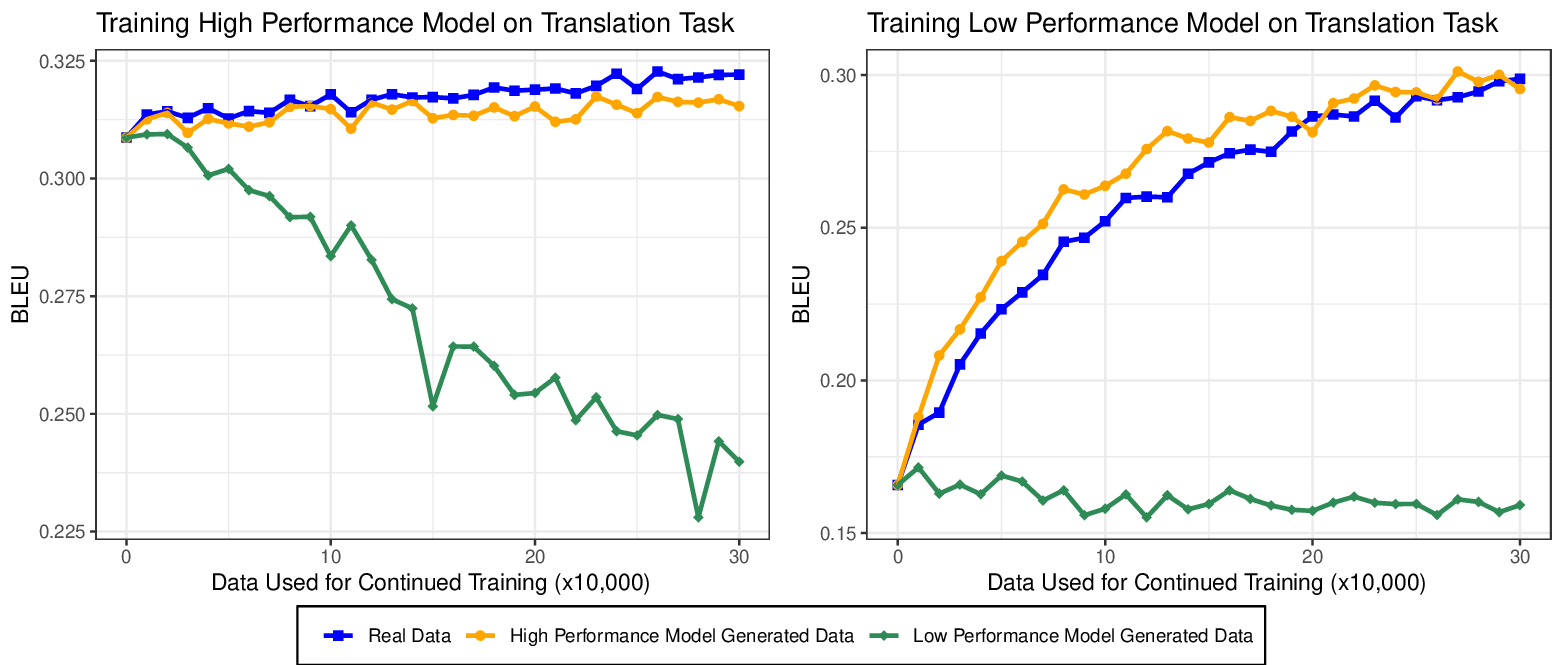}}
 \caption{Performance of Regurgitative Training Transformer Models (two plots have different $y$-axis scales for better readability) \label{fig:translation_res}}
\end{figure}

These two baseline models, corresponding to different performance levels, are then used to evaluate the effects of regurgitative training. We randomly sample a total of 300,000 data instances (outside of the training data of both the low- and high-performance baseline models) designated for regurgitative training. In batches of 10,000 data instances, we continue training both the low-performance baseline model and the high-performance baseline model with (i) real human-generated data, (ii) data generated by the low-performance model, and (iii) data generated by the high-performance model. After each batch of training, we evaluate all models' performances on the same testing data of 50,000 instances. The results are presented in Figure \ref{fig:translation_res}.

\begin{figure}[!tbh]
 \centering
  {\includegraphics[width=0.6\textwidth]{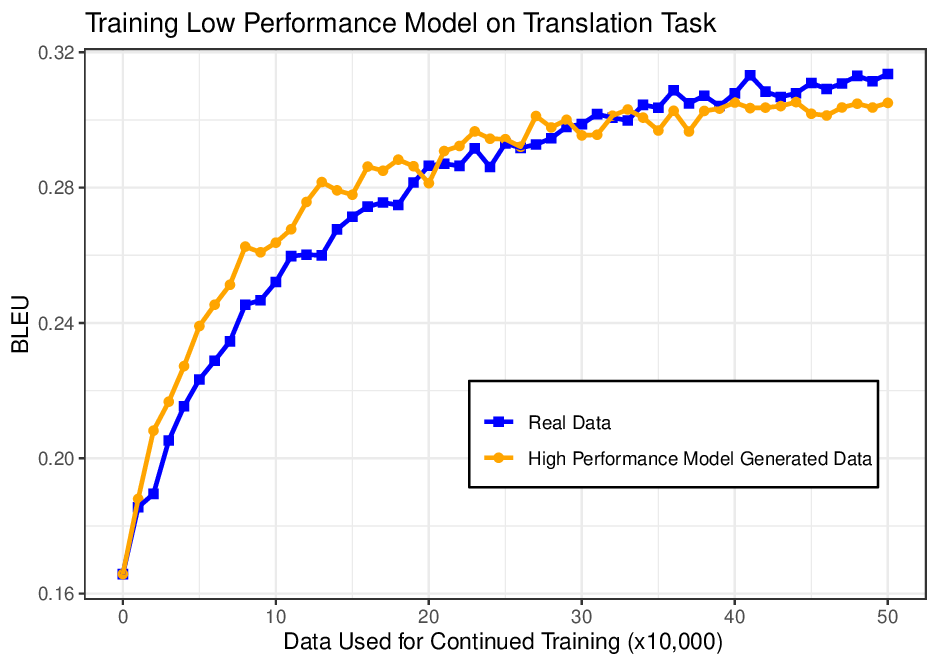}}
 \caption{Performance of Regurgitative Training Low-Performance Model (Augmented Data Size) \label{fig:translation_main_more}}
\end{figure}

For both low-performance and high-performance baseline models, regurgitative training with data generated from the low-performance model clearly underperforms training with real data. The same is true for regurgitative training of high-performance model with data generated by itself, though the performance gap is fairly small. Curiously, regurgitative training of low-performance model with data generated by high-performance model actually \textit{outperforms} training with real data for the first 19 batches (i.e., top two lines in the right plot). To understand whether this is a sustainable performance advantage, we sample more data to carry out another 20 batches of regurgitative training in this case. The results, as seen in Figure \ref{fig:translation_main_more}, show that regurgitative training performance starts to plateau around 30 batches, and underperforms training with real data thereafter.

The above results further demonstrate the performance cost of regurgitative training, even when businesses create and train their models from scratch. Consistent with our observations under regurgitative fine-tuning, regurgitative training with data from a more capable model is better than those from a less capable model -- training with data generated from the low-performance model clearly harms performance. Regurgitative training with the more capable high-performance model can match or even surpass the performance of training with real data, but such advantages usually fade away as the size of regurgitative training data grows.

\begin{figure}[!tbh]
 \centering
  {\includegraphics[width=\textwidth]{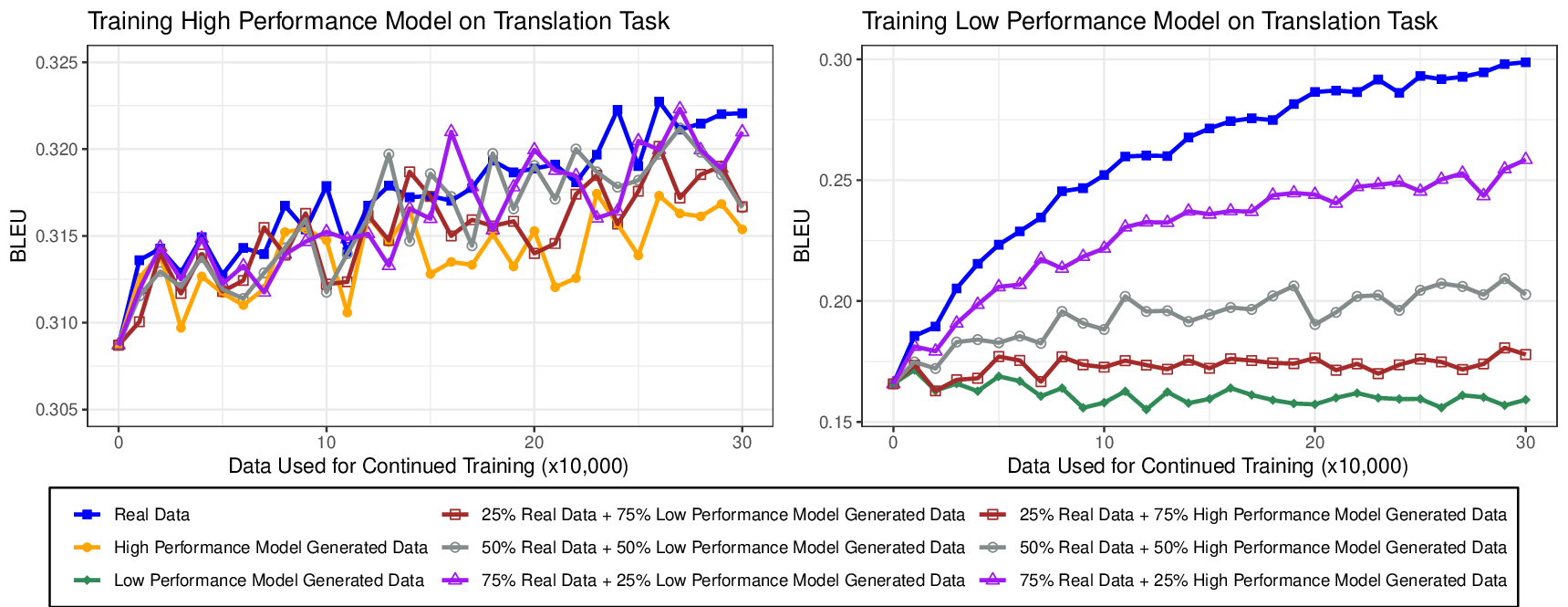}}
 \caption{Performance of Regurgitative Training Transformer Models with Different Proportions of Real Data (two plots have different $y$-axis scales for better readability) \label{fig:translation_res1}}
\end{figure}

In reality, it is plausible that a mixture of both model-generated data and real data are used for training. We therefore carry out another set of experiments to check how the proportion of model-generated data in the mix affects model performance. We simulate five scenarios, where the proportion of model-generated data is 100\%, 75\%, 50\%, 25\%, and 0\% respectively, and the rest of each mixture consists of real data. Naturally, the scenarios with 100\% and 0\% of model-generated data are the same as training purely with model-generated data or purely with real data. For clarity, we focus on training each baseline model with data generated by itself (mixed with different proportions of real data). Other experiment settings are the same as before, and the results are shown in Figure \ref{fig:translation_res1}.

In the case of high-performance baseline (i.e., left plot), because the performance gap between regurgitative training and training with real data is relatively small to begin with, the pattern is obfuscated by local performance fluctuations. Nonetheless, having a higher proportion of real data still generally leads to better performance. In contrast, the pattern becomes much clearer in the case of low-performance baseline (i.e., right plot). We can see that even a small amount of model-generated data is enough to deteriorate learning. As a higher proportion of model-generated data is used, the model's performance continues to deteriorate.

Finally, we note that because the high-/low-performance baseline models are initially trained with different volumes of data, the data added to the two models during regurgitative training naturally amount to different proportions of their initial training data -- specifically, each batch of 10,000 data corresponds to 20\% of low-performance model's training data and 2\% of high-performance model's training data. As a robustness check, we repeat regurgitative training of the two models by adding a \textit{fixed percentage} of each model's training data, i.e., 10\% per batch. The results, reported in Appendix \ref{ap:percentage}, remain highly consistent with our main findings above.

\subsection{Replication: Question Answering Task} \label{sec:experiments_self_train_QA}
In addition to machine translation, we also conduct a replication study with another common generative language task -- Question Answering (Q\&A). We use the Stanford Question Answering Dataset (SQuAD) \citep{rajpurkar2016squad}, which is a widely used benchmarking dataset for developing and testing Q\&A methods. SQuAD is a reading comprehension dataset composed of questions created by crowd-workers based on a collection of Wikipedia articles, with answers being segments of texts from the corresponding passages in the articles. The dataset includes 87,599 entries in the training dataset (constructed from 442 articles) and 10,570 entries in the development dataset (constructed from 48 other articles) which we use as testing data.

Instead of end-to-end training (as was done in the previous section), here we use a pre-trained BERT model \citep[\texttt{bert-base-cased},][]{devlin2018bert} to extract word embeddings, which are fed into a feedforward neural network model for Q\&A. Doing so allows us to evaluate regurgitative training under yet another widely adopted strategy for training generative language models (i.e., leveraging pre-trained representation models).\footnote{This strategy is often also referred to as ``fine-tuning" a pre-trained model. We refrain from using this term here in order to avoid confusion with our fine-tuning experiments in Section \ref{sec:experiments_finetune}, which are performed on top of existing LLMs (rather than a BERT-like representation model).} We follow \cite{rajpurkar2016squad} to evaluate Q\&A performance with two metrics: \textit{Exact Match} and \textit{F-1 score}. Exact match measures the percentage of predicted answers that match the ground-truth answers exactly, and F-1 score measures the overlap between the predicted answers and the ground-truth answers by treating both predictions and ground-truths as bags of tokens. We calculate the average F-1 score over all questions in the testing data. 

Following the same procedure as in the previous section, we use increasing amounts of real data to train baseline models in order to identify a low-performance model and a high-performance model (see Appendix \ref{ap:QA_real} for detailed results). The low-performance model is trained on all entries from 40 articles and achieves 70.68\% exact match rate and 80.38\% average F-1 score, whereas the high-performance model is trained on all entries from 200 articles and achieves 78.47\% exact match rate and 86.54\% average F-1 score.

\begin{figure}[!tbh]
 \centering
  {\includegraphics[width=0.9\textwidth]{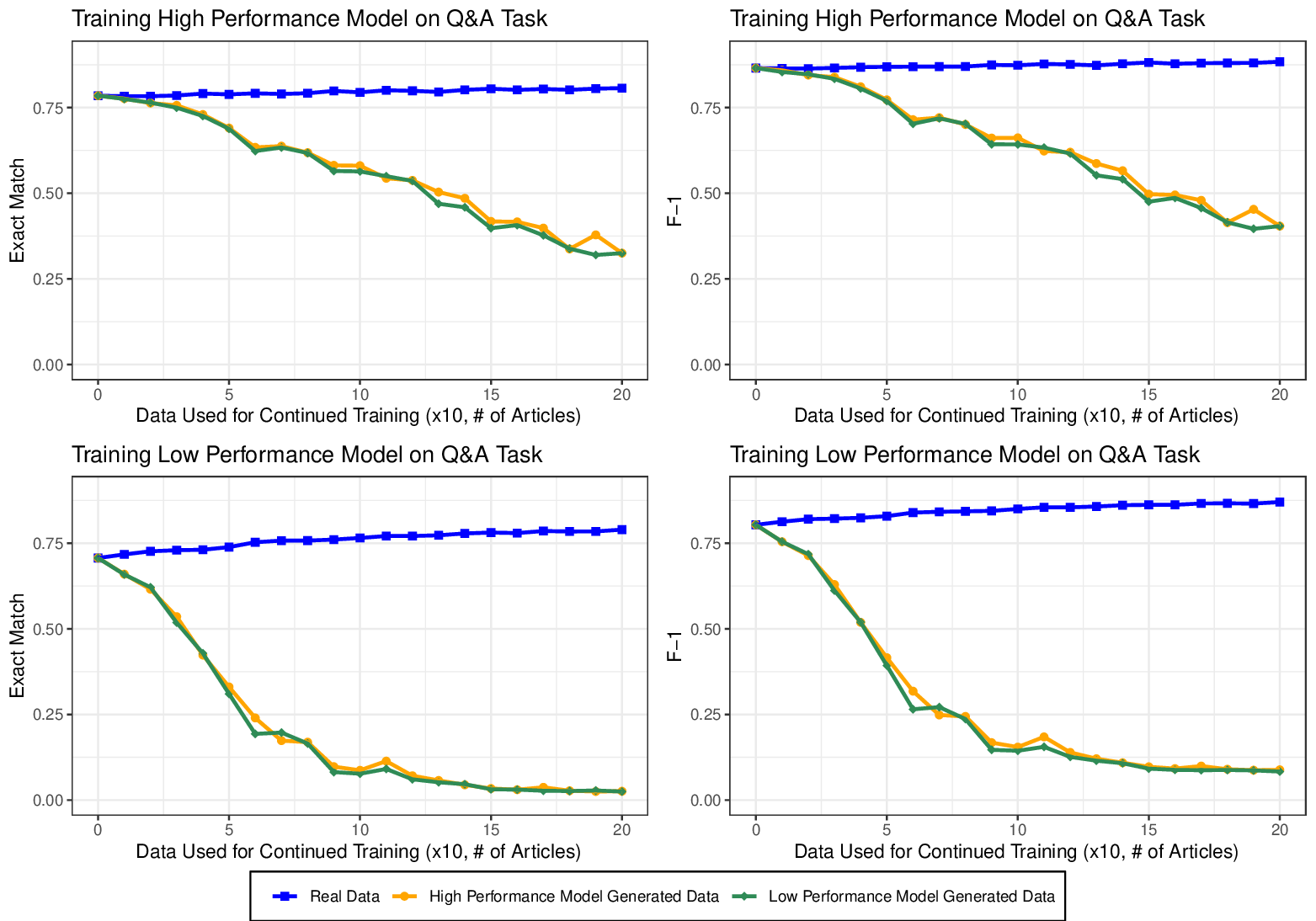}}
 \caption{Performance of Regurgitative Training Transformer Models on Q\&A Tasks \label{fig:QA_res}}
\end{figure}

Next we use these two baseline models for regurgitative training on entries from the remaining 242 articles (not used in training for the two baseline models). In batches of 10 articles, we continue training both the low- and high-performance baselines with (i) real human-generated data, (ii) data generated by the low-performance model, and (iii) data generated by the high-performance model, for a total of 20 batches. After each batch of training, we evaluate all models' performances on the same testing dataset provided by SQuAD. The results are included in Figure \ref{fig:QA_res}, where the first row shows performance of regurgitative training the high-performance baseline model and the second row shows performance of regurgitative training the low-performance baseline model. 

We again observe that regurgitative training with model-generated data negatively affects Q\&A performance, compared to training with real data. Different from the translation task, Q\&A regurgitative training with data generated by the high-performance model does not improve performance (and certainly does not outperform training with real data), even though it still weakly outperforms regurgitative training with low-performance model generated data. In other words, the peril of regurgitative training is not limited to translation task and is even more severe in Q\&A task.

\section{Understanding Performance Loss from Regurgitative Training} \label{sec:mechanism}
Why does regurgitative training hurt performance compared to training with real data? In this section, we offer some preliminary evidence into the underlying mechanisms. Using the translation task as an example, we focus on characterizing the differences between LLM-generated training data and real data, and discuss how these differences may impact the performance. Keeping in mind that theoretical / mathematical understandings of LLMs (and complex deep neural networks in general) still remain largely out of reach, the following explorations are empirical in nature.

The first mechanism is \textit{error} -- LLMs are not perfect and data generated by them can contain more errors than real data. New models trained on these error-prone data can therefore have inferior performance. This is also the mechanism identified and studied in prior work \citep{shumailov2023curse}. We test this mechanism in the fine-tuning setting of Section \ref{sec:experiments_finetune} with the 5,000 data points used for regurgitative training. Recall that, with access to ground-truths for these data points, we have already calculated the BLEU scores of translations generated by GPT-3.5, GPT-4, and LLAMA2. We have confirmed that translations generated by GPT-4 have a slightly higher BLEU score than those generated by GPT-3.5, and both clearly have higher BLEU scores than LLAMA2-generated data. This aligns well with the testing performance of the corresponding fine-tuned LLMs (Figure \ref{fig:gpt_performance}).

Although BLEU is widely used to measure translation quality, it also has an important limitation that it does not explicitly account for the semantic meaning of words. A translation that uses different words than those in the ground-truths will have a low BLEU score even if it is semantically correct. In other words, having a lower BLEU score does not necessarily mean that the translation is more erroneous. In light of this, we construct two new measures, both aiming at quantifying the semantic differences of LLM-generated data vs. real data. We take each set of training data (generated by one of the LLMs or human) and perform several pre-processing steps, including (i) lower-casing, (ii) removing punctuation, (iii) removing stopwords, and (iv) lemmatization (i.e., reducing a word to its stem form). These pre-processing steps allow us to focus only on the substantive content of each translation.\footnote{For example, two different translations ``Tomorrow will be raining!" and ``Tomorrow will rain." will both become ``tomorrow rain" after pre-processing, as it lower-cases both sentences, removes punctuation, removes stopwords ``will" and ``be", and reduces ``raining" to its stem ``rain".} The first metric is computed as the average cosine similarity between the embeddings of LLM-generated and ground-truth translations, where the embeddings are obtained from the Sentence Transformer model \citep{reimers2019sentence}. After pre-processing, a smaller cosine similarity implies greater semantic discrepancies of LLM translations from the ground-truths, which is indicative of translation errors. The second metric counts the number of word tokens in a ground-truth translation that satisfy two conditions: (1) they do not show up in the corresponding LLM translation \textit{and} (2) even their synonyms \citep[retrieved based on WordNet,][]{miller1995wordnet} do not show up in the LLM translation. These non-synonymous deviations likely represent words mistranslated by LLM. Results of these two metrics are reported in the second and third rows of Table \ref{tab:mechanism_error}. We see that the two sets of GPT-generated data have higher semantic similarities with ground-truths and lower non-synonym deviations than LLAMA2, again supporting the mechanism that translation errors are partially responsible for the performance reduction of regurgitative training.

\begin{table}[!tbh]
    \centering 
    \caption{Metrics of Translation Errors and Comparison Results} \label{tab:mechanism_error}
    \begin{tabular}{c|c|c|c}
    \hline
         &  GPT-3.5 vs. Real  &  GPT-4 vs. Real  &  LLAMA2 vs. Real \\
    \hline
    Average Cosine Similarity  &  0.8047  &  0.8059  &  0.7506 \\ 
    Total \# of Non-Synonymous Deviations  &  21836  &  21619  &  27962 \\
    \hline
    \end{tabular}    
\end{table}

Beyond errors, we also test a different mechanism related to \textit{lexical diversity}. Several recent work suggest that LLM-generated content appears to be more homogeneous than human-generated content \citep{doshi2023generative,anderson2024homogenization,zhou2024generative}. We suspect that regurgitative training with less diverse LLM-generated data may hinder the model's ability to generalize and result in lower testing performance. We quantify lexical diversity with two metrics. The first is a straightforward count of the total number of unique word tokens in ground-truth or LLM translations. The second adopts the self-BLEU metric proposed by \cite{zhu2018texygen}. Self-BLEU is the BLEU score of a given text against all other texts in a corpus. Because BLEU captures lexical similarity, self-BLEU accordingly reflects how similar a text is with the rest of the corpus (higher self-BLEU implies \textit{lower} diversity). For LLM-generated translations, errors may artificially decrease self-BLEU without meaningfully increase lexical diversity. We therefore remove the previously mentioned non-synonymous deviations (as approximation of errors) from LLM translations. We then average self-BLEU over the 5,000 training data points. Results of both metrics are reported in Table \ref{tab:mechanism_diversity}. Ground-truth translations consistently use more unique tokens and have significantly lower average self-BLEU than LLM translations ($p < 0.001$). GPT translations use a bit more unique tokens than LLAMA2 and the three LLMs have similar average self-BLEU (their self-BLEU differences are not statistically significant, $p > 0.05$). 

\begin{table}[!tbh]
    \centering 
    \caption{Metrics of Lexical Diversity and Comparison Results} \label{tab:mechanism_diversity}
    \begin{tabular}{c|c|c|c|c}
    \hline
         &  Real  &  GPT-3.5  &  GPT-4  &  LLAMA2  \\
    \hline
    Total \# of Unique Tokens  &  14604  &  13690  &  13731  &  13081 \\ 
    Average Self-BLEU  &  0.1048  &  0.1154  &  0.1154  &  0.1126 \\
    \hline
    \end{tabular}    
\end{table}

Given the black-box nature of LLMs, we acknowledge that the exact process through which errors or lack of lexical diversity in training data affect model performance remains unclear. Nonetheless, these explorations provide plausible explanations for the negative performance impact of regurgitative training. More importantly, they naturally give rise to potential strategies to mitigate performance loss due to regurgitative training. We investigate a few different strategies in the next section.

\section{Mitigating Performance Loss from Regurgitative Training} \label{sec:mitigation}
In this section, we propose and test a few strategies to mitigate the adverse performance impact of regurgitative training. Designing effective mitigation strategies requires first understanding the mechanisms of the adverse effects. Our explorations in the previous section provide suggestive evidence that errors and lack of lexical diversity may both be at play. Accordingly, we design three mitigation strategies to address one or both of these mechanisms:
\begin{itemize}
    \item Strategy 1 relies on quality quantification to gauge the likelihood of errors in synthetic data, and prioritize the use of data with high quality (i.e., low error likelihood) in regurgitative training;
    \item Strategy 2 seeks to enhance lexical diversity by mixing together synthetic data generated by different LLMs in regurgitative training;
    \item Strategy 3 builds an AI detection model to differentiate between synthetic vs. real data, and prioritize the use of synthetic data that most resemble real data for regurgitative training. As a competent AI detector may pick up on both errors and lexical diversity as predictive features, this strategy is designed to address both issues.
\end{itemize}

Details of each strategy and the corresponding evaluations on the translation task are discussed in the rest of this section. The first quality-based mitigation strategy is also naturally applicable on the Q\&A task, which we will demonstrate as part of Section \ref{sec:mitigation_selfbuilt}. However, it is worth noting up front that the goal of mitigation is not to completely close the gap from the performance of training with real data -- this may not be realistic in the short term. Instead, the goal is to use LLM-generated synthetic data in a more careful manner to \textit{reduce} performance loss.\footnote{One might suggest the best strategy to reduce performance loss is not to use any synthetic data at all. However, as we discussed in Section \ref{sec:introduction}, real human-generated data alone may not be sufficient to train next-generation LLMs.}

\subsection{Mitigation Strategy based on Quality Quantification} \label{sec:mitigation_quality}
The first strategy is to identify a method to assess the quality of synthetic data, and subsequently select higher-quality data for regurgitative training. This requires defining a metric that accurately measures, or at least correlates with, data quality specific to the task at hand. One such metric, commonly used in classification contexts, is prediction confidence score. Higher prediction confidence scores usually correlate with greater probability of correct predictions, and the semi-supervised learning literature routinely uses prediction confidence as a quality metric \citep[e.g.,][]{scudder1965probability}. Because modern LLMs generate content by autoregressively predicting the next token, it is viable to also adopt prediction confidence, calculated based on predicted probabilities over the vocabulary, to quantify the quality of LLM-generated data. However, a practical obstacle is that when using third-party LLMs, prediction probabilities may not always be available. Therefore, we devise an alternative quality metric to guide the quality-based mitigation in the setting of LLM fine-tuning, assuming prediction probabilities are unavailable (Section \ref{sec:mitigation_finetune}). We also demonstrate the same mitigation strategy with transformers trained from scratch, assuming prediction probabilities are fully available (Section \ref{sec:mitigation_selfbuilt}).

\subsubsection{Evaluation in Fine-Tuning Setting.} \label{sec:mitigation_finetune}
In translation task, in the absence of raw prediction probabilities, the BLEU score can be used as another metric to gauge data quality. We propose to train a supervised learning model to predict the BLEU score of a LLM-generated translation. To train such a BLEU prediction model, we randomly sample 150,000 German-English sentence pairs (not previously used in Section \ref{sec:experiments_finetune}) and obtain the translations generated by GPT-3.5, GPT-4, and LLAMA2. For each pair of German sentence and LLM translation, we compute the BLEU score using the ground-truth translation as the reference. These LLM-generated translation pairs, along with their BLEU scores, form the labeled dataset for training the BLEU prediction model.

The labeled dataset is randomly split into 80\% for training and 20\% for testing. Each instance of the labeled dataset is structured as $input = (g_1, g_2,\ldots, g_M, [SEP], e_1, e_2, \ldots, e_N, [SEP]), label = BLEU$, where $g_i$ represents tokens in German sentences, $e_j$ represents tokens in English translations, and $[SEP]$ denotes the special separation token. Similar to an approach used in \cite{chowdhury2021ensemble}, we derive embedding of the entire input sequence from multilingual BERT (with the \texttt{bert-base-multilingual-uncased} pre-trained model), which is then used as input features to eight different supervised learning techniques for BLEU prediction. We train a separate BLEU prediction model for each of the three LLMs, and the testing performance of these BLEU prediction models are summarized in Appendix \ref{ap:quality_bleu_prediction}. We find that the Bayesian Ridge technique exhibits relatively superior BLEU prediction performance (achieving lower MSE and MAE values). 

\begin{figure}[!tbh]
 \centering
  {\includegraphics[width=\textwidth]{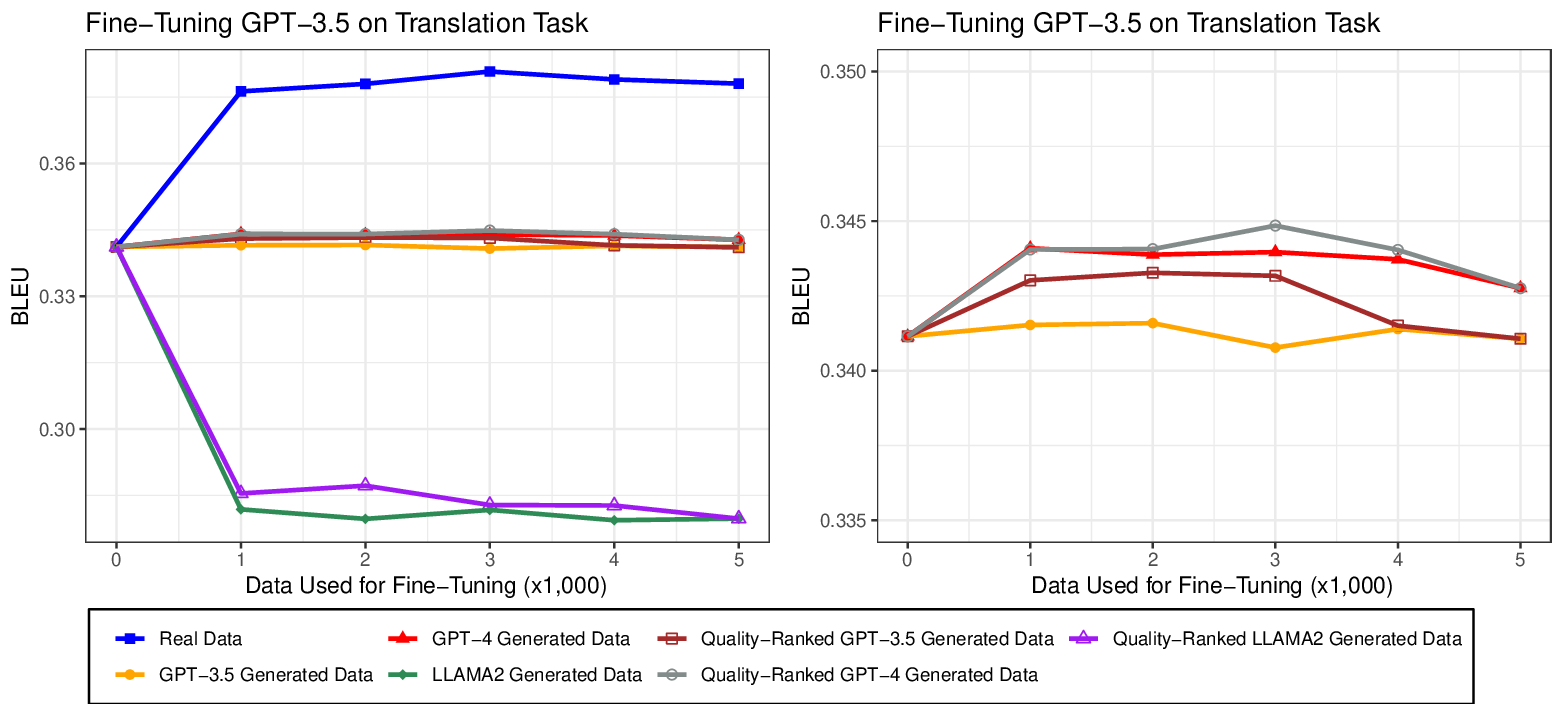}}
 \caption{Quality-Based Mitigation Strategy: Results on LLM Fine-Tuning (plot on the right zooms in on GPT-related results for better readability) \label{fig:finetune_quality}}
\end{figure}

Using the best-performing BLEU prediction model for each LLM, we predict the BLEU scores of the 5,000 LLM-generated translations previously used for regurgitative fine-tuning. Next, we rank the LLM-generated translations by their predicted BLEU scores, from high to low, then proceed to fine-tune the baseline GPT-3.5 model in batches of 1,000 data instances and evaluate the resulting performance. The batch-wise fine-tuning procedure and the testing data partition used for performance evaluation are exactly the same as in Section \ref{sec:experiments_finetune}. We present the results in Figure \ref{fig:finetune_quality}. Please note that, because BLEU scores from regurgitative training with GPT models have much smaller variations than those from real data or LLAMA2 generated data, we also add a plot on the right side of the Figure to zoom in on GPT-related results.

We can see that regurgitative training using quality-ranked data shows some improvements compared to using the corresponding LLM-generated data without quality consideration, thereby supporting the utility of quality-based mitigation strategy. However, across the three LLMs we have tested, the magnitudes of performance improvement are all rather small and still far from reaching the performance level of training with real data.

\subsubsection{Evaluation in Training-from-Scratch Setting.} \label{sec:mitigation_selfbuilt}
When businesses build their own language models, as described in Section \ref{sec:experiments_self_train}, a naturally available metric for evaluating data quality is the prediction confidence score, typically calculated based on class probability predictions. In the transformer architecture, these probabilities are the outputs of the softmax layer. Rather than using the highest (i.e., top-1) predicted probability to measure data quality, which has been shown to lead to overconfidence \citep{zhang2021knowing,lyu2020you}, we follow \cite{fomicheva2020unsupervised} and use the entropy of the probability distribution over the entire vocabulary. Mathematically, given a translation with $T$ tokens, we calculate the entropy of probability distribution over vocabulary $V$ for each generated token $t \in \{1, \ldots, T\}$, then average the token-level entropy scores to form an overall translation-level entropy score:
\begin{equation}
\label{eq:translation_confidence}
\text{Translation Entropy}  =-\frac{1}{T}\sum_{t=1}^{T}\sum_{v\in V}p(y_t^v)\log{p(y_t^v)})
\end{equation}
where $p(y_t^v)$ denotes the predicted probability of candidate token $v \in V$ at position $t$. A lower entropy score indicates a more confident translation.

\begin{figure}[!tbh]
 \centering
  {\includegraphics[width=\textwidth]{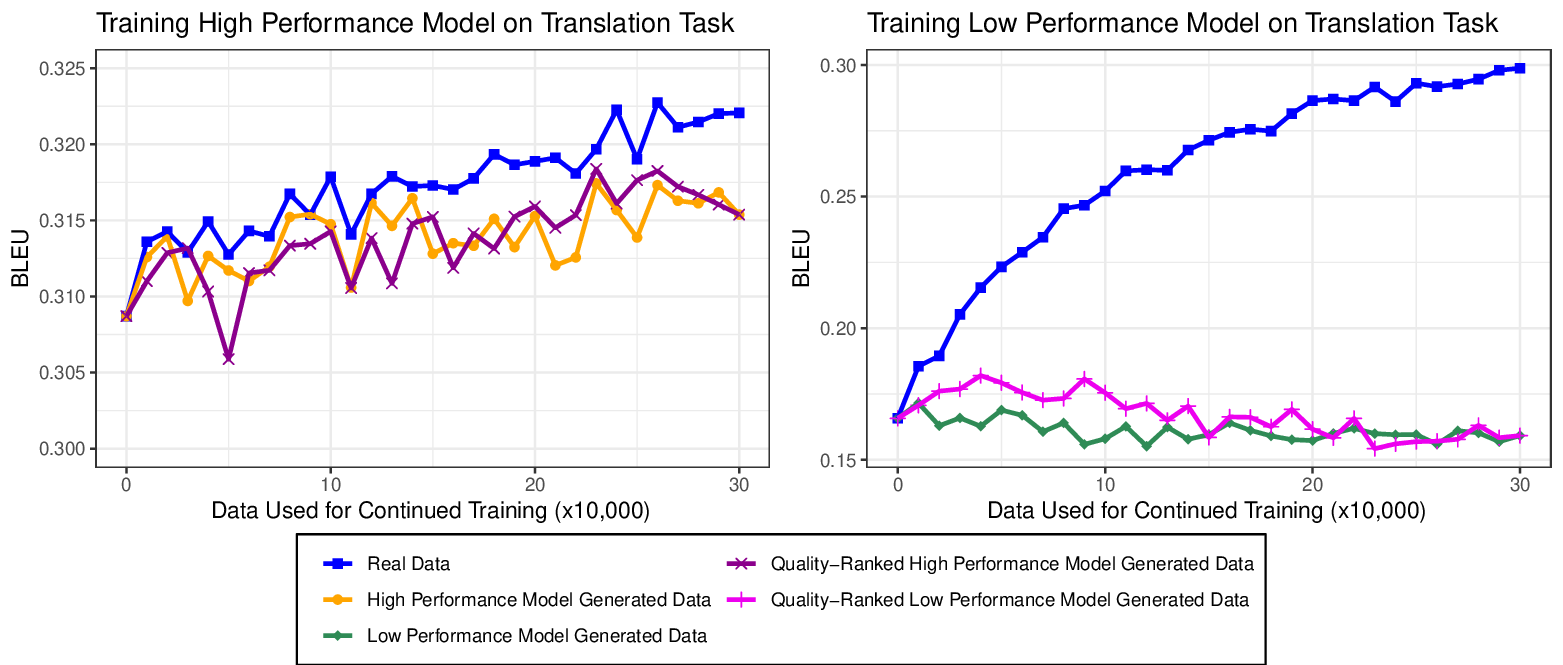}}
 \caption{Quality-Based Mitigation Strategy: Results on Transformer Models Trained from Scratch (two plots have different $y$-axis scales for better readability) \label{fig:sort_translation}}
\end{figure}

We carry out regurgitative training by incorporating model-generated data ranked by their translation entropy scores, from low to high (equivalent to ranking data based on translation confidence, from high to low). Recall that we have trained both a low-performance baseline model and a high-performance baseline model and, accordingly, the quality-based regurgitative training is done for both models. The rest of the experiment settings, including the progressive training and evaluation procedure, are kept exactly the same as in Section \ref{sec:experiments_self_train}. The results are displayed in Figure \ref{fig:sort_translation}. Note that, unlike Figure \ref{fig:translation_res}, this figure does not contain results from adding data generated by high-performance model to train the low-performance model or vice versa. This is because it may not be reasonable to expect probability predictions to be readily available from a different model other than the baseline.

When the baseline model has low performance (right plot), we find that regurgitative training with quality-ranked data can mitigate performance loss to some extent, similar to what has been observed under the fine-tuning setting. The performance gain is especially evident when a relatively small amount of data (roughly one third of all model-generated data) are added. However, when the baseline model has high performance (left plot), we do not observe consistent performance benefits of the quality-based mitigation strategy. This is likely because the performance gap from training with real data is already small and performance fluctuations (e.g., due to randomness in data) may obfuscate a clear pattern of the quality-based mitigation strategy in this case.

Finally, we also apply the quality-based mitigation strategy on the Q\&A task. For a given question, we analogously derive an entropy score for a model-generated answer to reflect the uncertainty of the probability distribution over all possible answers. For Q\&A task, the transformer model generates a candidate answer by predicting the positions of a start token and an end token in a given passage that potentially contains the answer \citep{rajpurkar2016squad}. The start / end tokens then jointly determine the answer text. Formally, the probability score of a candidate answer is defined as the product of the probabilities associated with the start and end tokens after a softmax transformation, and the entropy score can be computed as 
\begin{equation}
\label{eq:QA_confidence}
\text{Answer Entropy} = -\sum_{a\in A} p(y_a) \log \left \{ p(y_a)  \right \}
\end{equation}
where $a\in A$ is a candidate answer and $p(y_a)=\frac{p(y_a^{start}) p(y_a^{end})}{\sum_{a\in A}p(y_a^{start}) p(y_a^{end})}$. Same as before, a lower entropy corresponds to a more confident model-generated answer.

\begin{figure}[!tbh]
 \centering
  {\includegraphics[width=0.9\textwidth]{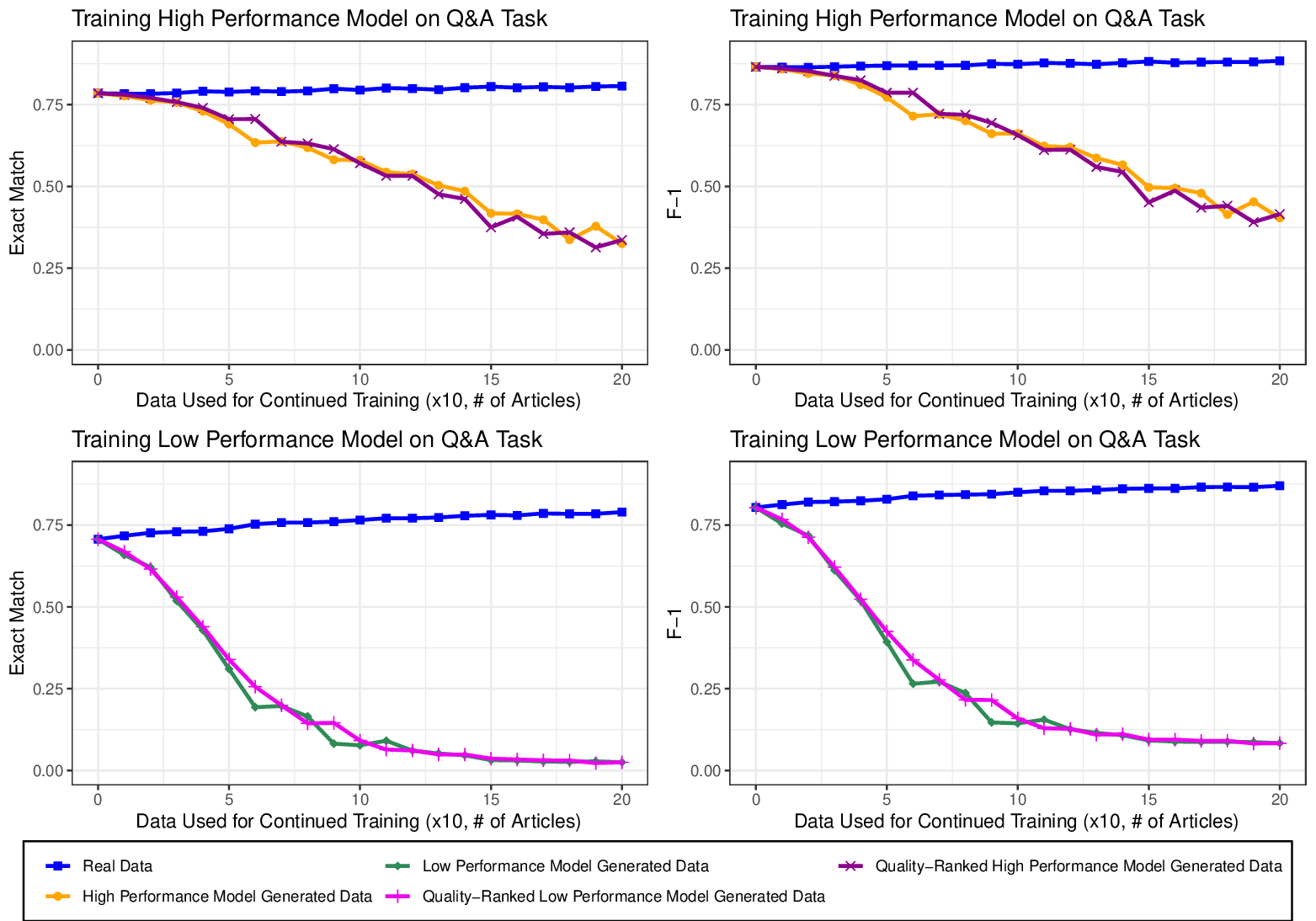}}
 \caption{Quality-Based Mitigation Strategy: Results on Q\&A Task \label{fig:sort_QA}}
\end{figure}

We then follow the same settings in Section \ref{sec:experiments_self_train_QA} to conduct regurgitative training on both low-performance and high-performance baseline models, taking into account answer entropy. Recall that each article contains multiple training instances (question-answer pairs). Within an article, we rank answers by their entropy scores, from low to high. Across articles, we prioritize articles based on the lowest entropy value among their constituent answers. The results are shown in Figure \ref{fig:sort_QA}.

We again observe some performance gains of regurgitative training with quality-ranked data, but the improvements tend to be fairly small and dissipate after a sufficient amount of data are added. We repeat this set of experiments but instead prioritize articles based on the \textit{average} entropy among their constituent answers, and the results remain qualitatively unchanged (Appendix \ref{ap:qa_avg}). To summarize, while the experiments in this section generally support the potential of the quality-based mitigation, we note that the performance gap from training with real data remains substantial. Put differently, the value of real data in training language models cannot be substituted by quality-aware regurgitative training.

\subsection{Mitigation Strategy based on Data Mixture} \label{sec:mitigation_mixture}
As discussed in Section \ref{sec:mechanism}, a lack of lexical diversity in LLM-generated content may also contribute to the performance loss of regurgitative training. This observation prompts us to explore a mitigation strategy aimed at enhancing lexical diversity within LLM-generated data. Specifically, we propose mixing together data generated from multiple LLMs into the regurgitative training process. For example, LLMs developed by different companies, each potentially trained on somewhat distinct datasets, may consequently produce data with unique characteristics and nuances. Combining data from different ``breeds" of LLMs can therefore introduce greater variability than relying on a single LLM.

We first test this data mixture strategy under LLM fine-tuning setting. With three LLMs, there are 3 possible mixture configurations: (i) mixing GPT-3.5 with GPT-4, (ii) mixing GPT-3.5 with LLAMA2, and (iii) mixing GPT-4 with LLAMA2. We consider configurations (i) and (iii) in particular. Configuration (iii) mixes two top-of-the-line LLMs developed by different companies and, based on our explorations in Section \ref{sec:mechanism}, they exhibit different lexical diversity. In other words, this configuration is a more direct evaluation of the proposed strategy. However, as we have shown before, LLAMA2-generated data have higher translation error rates than GPT-generated data, so the performance outcomes of this configuration may not be solely attributed to changes in lexical diversity levels. In comparison, configuration (i) is less confounded because the two GPT models have similar translation quality, although their mixture also bring in less additional lexical variations. Taken together, we present results from both configurations to offer a more comprehensive evaluation of the data mixture strategy.

\begin{figure}[!tbh]
 \centering
  {\includegraphics[width=0.7\textwidth]{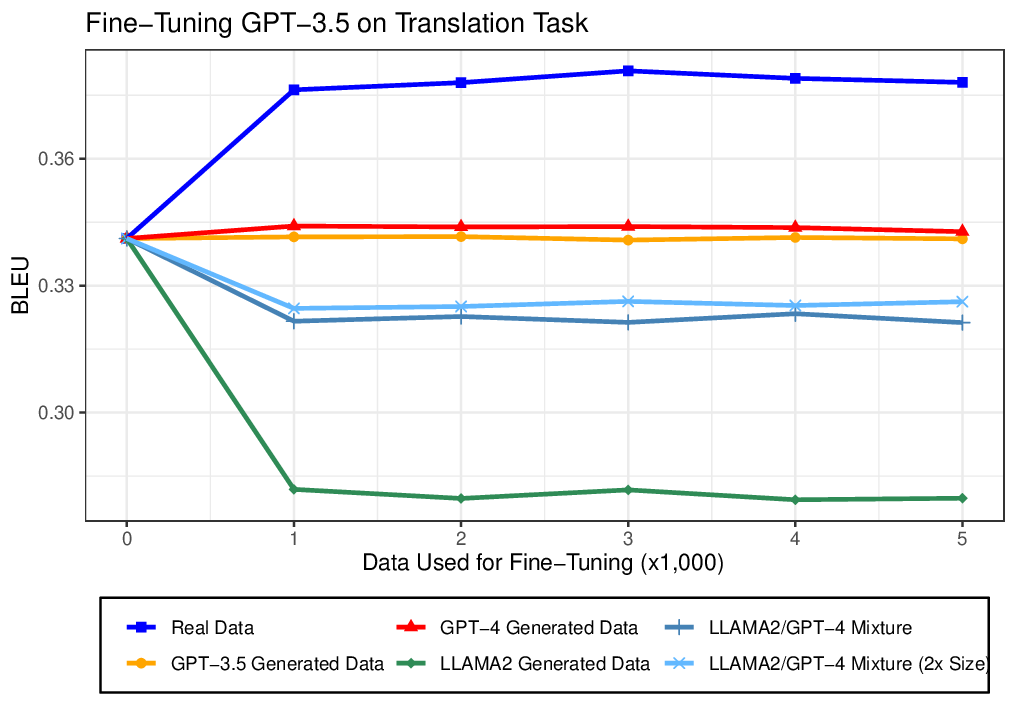}}  
 \caption{Data Mixture-Based Mitigation Strategy: GPT-4 / LLAMA2 Mixture Results on LLM Fine-Tuning \label{fig:gpt_lgmix_mix}}
\end{figure}

\begin{figure}[!tbh]
 \centering
  {\includegraphics[width=\textwidth]{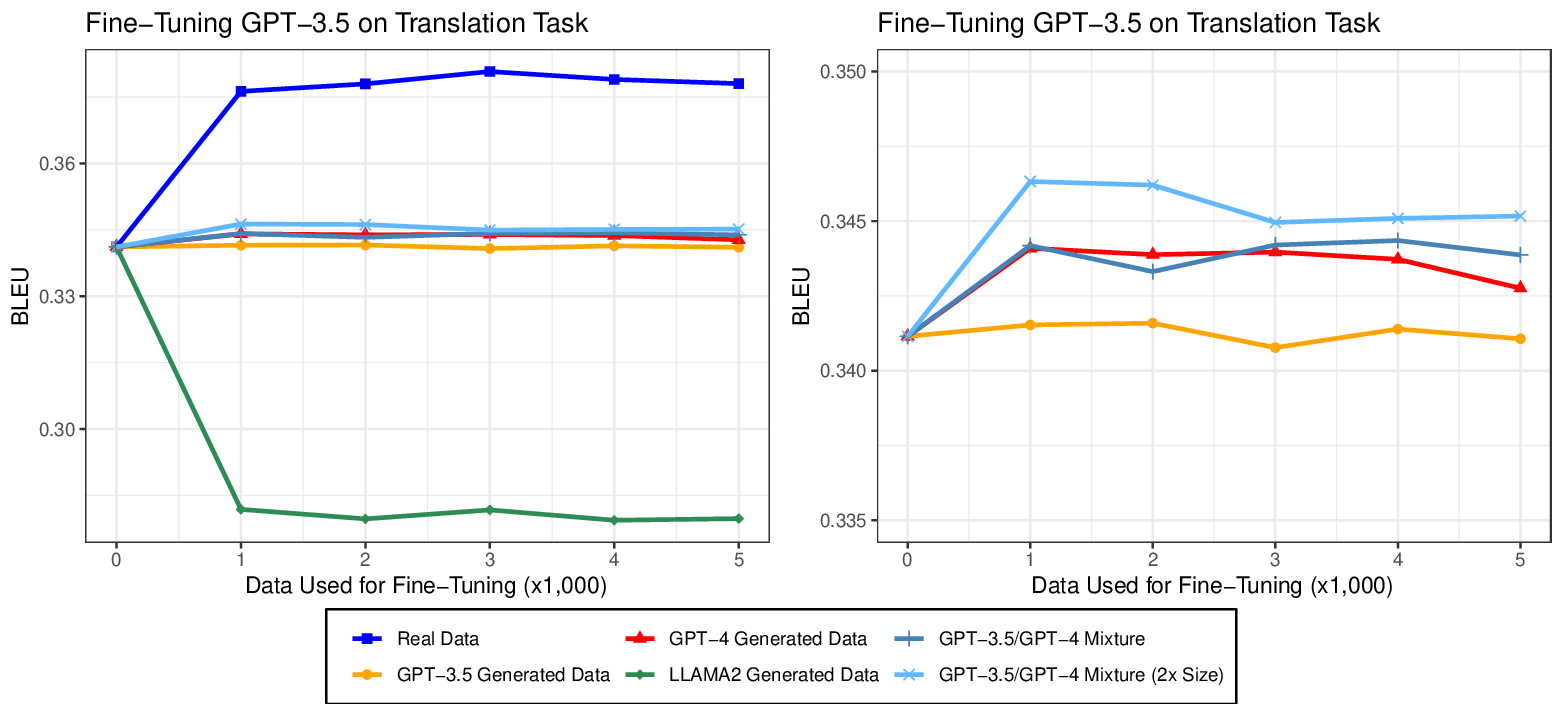}}  
 \caption{Data Mixture-Based Mitigation Strategy: GPT-3.5 / GPT-4 Mixture Results on LLM Fine-Tuning (plot on the right zooms in on GPT-related results for better readability) \label{fig:gpt_gpt34mix_mix}}
\end{figure}

Furthermore, for each batch of 1,000 regurgitative training instances, there are two ways to add the mixture data. First, we can add (randomly selected) 500 instances from one LLM and 500 from the other LLM, thereby keeping the total training batch size unchanged. Alternatively, we can add all 1,000 instances from both models, which amounts to a training batch size of 2,000. Both are reasonable from a practical perspective, and we report both sets of results. After each batch of regurgitative training, we evaluate translation performance on the same testing data used in Section \ref{sec:experiments_finetune}. Figure \ref{fig:gpt_lgmix_mix} shows results for GPT-4 / LLAMA2 mixture and Figure \ref{fig:gpt_gpt34mix_mix} shows results for the GPT-3.5 / GPT-4 mixture.

Interestingly, mixing data generated by GPT-4 and LLAMA2 results in performance levels that fall in between using the two constituent LLMs alone, but mixing data generated by GPT-3.5 and GPT-4 can match the performance of GPT-4 (the better-performing LLM of the two) and even slightly outperform it when twice the regurgitative training data are added. These results suggest that the effectiveness of the data mixture strategy is nuanced. When constituent LLMs differ both in terms of quality and lexical diversity, mixing their data together may not lead to better regurgitative training performance, because the potential benefit of greater lexical diversity is offset by having a more error-prone LLM in the mix. In the same vein, if two LLMs with similar quality yet different levels of lexical diversity can be identified, their mixture can indeed mitigate performance loss of regurgitative training to some extent.

\begin{figure}[!tbh]
 \centering
 {\includegraphics[width=\textwidth]{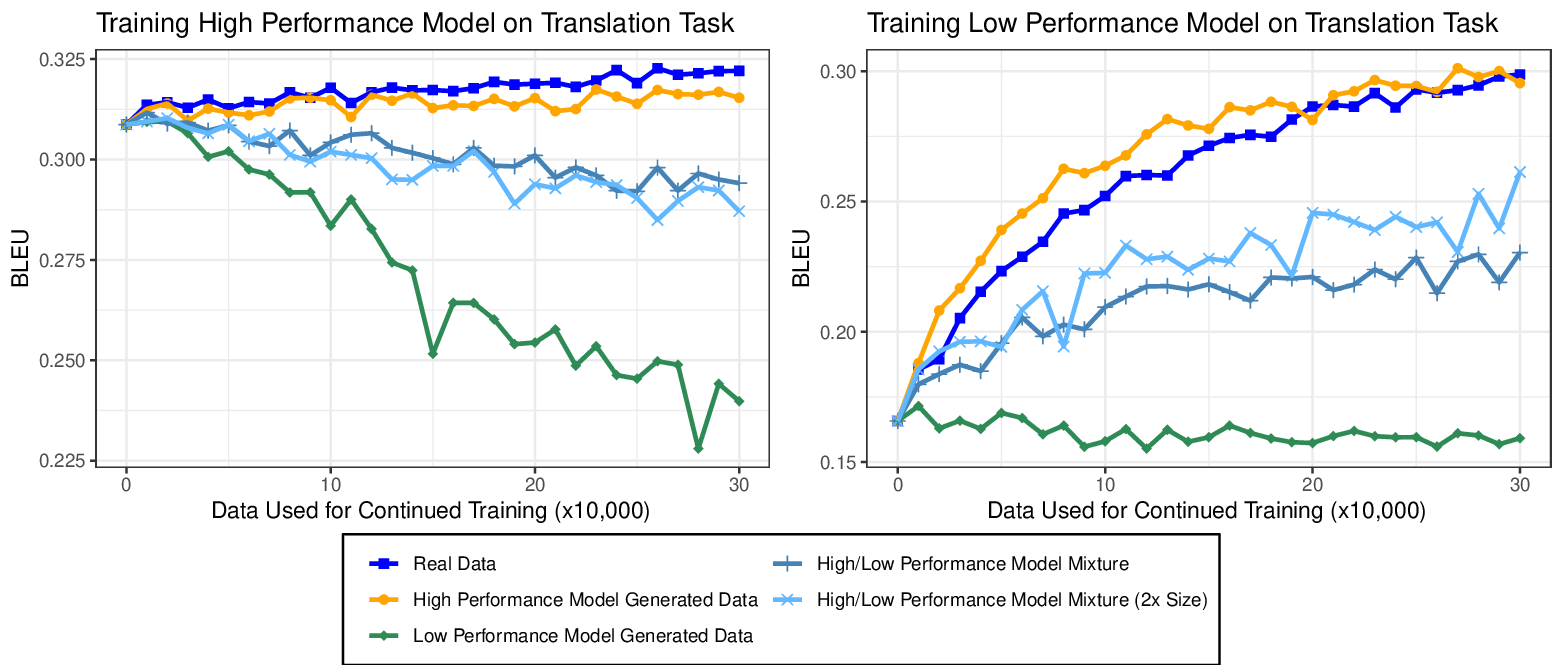}}  
 \caption{Data Mixture-Based Mitigation Strategy: High/Low Performance Model Mixture Results on Transformer Models Trained from Scratch (two plots have different $y$-axis scales for better readability) \label{fig:mitigation_mix_hl}}
\end{figure}

\begin{figure}[!tbh]
 \centering
 {\includegraphics[width=\textwidth]{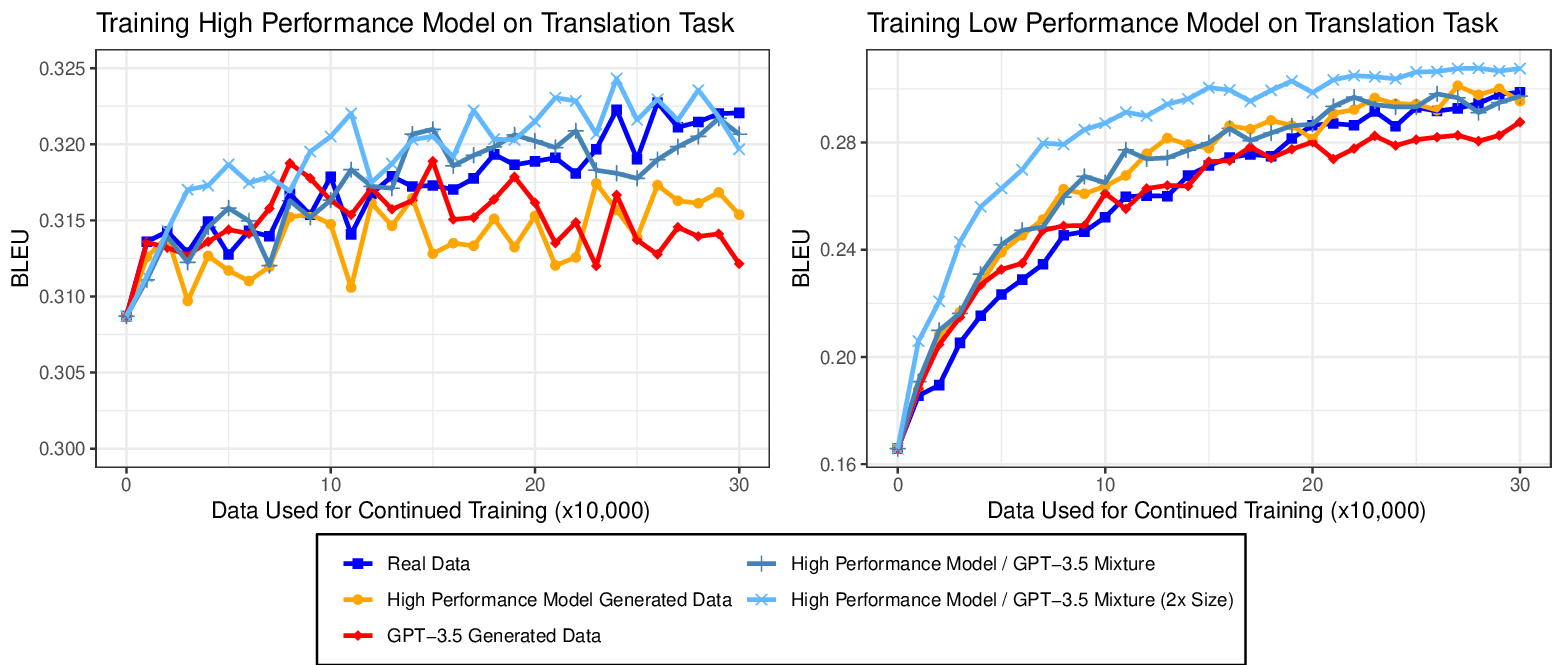}}  
 \caption{Data Mixture-Based Mitigation Strategy: High Performance Model / GPT-3.5 Mixture Results on Transformer Models Trained from Scratch (two plots have different $y$-axis scales for better readability) \label{fig:mitigation_mix_hg}}
\end{figure}

We also deploy the same strategy on transformer models trained from scratch. With two baseline models, there is only one mixture configuration, namely mixing data generated by the low-performance model and high-performance model. However, these two models naturally differ on both quality and lexical diversity (just like the GPT-4 / LLAMA2 mixture). Therefore, we also consider a mixture of data generated by GPT-3.5 and high-performance model (to mimic the GPT-4 / GPT-3.5 mixture configuration). Here, GPT-3.5 serves as a high-quality translation model that is also very different from the model we trained from scratch. The results are shown in Figures \ref{fig:mitigation_mix_hl} and \ref{fig:mitigation_mix_hg}.

We find highly consistent results as in the fine-tuning case. Mixing data generated by the two baseline models, which differ on quality, results in regurgitative training performance that falls between the two constituent models. Moreover, when training the high-performance baseline model with the data mixture (left plot of Figure \ref{fig:mitigation_mix_hl}), the performance follows a downward trend, likely because data generated by low-performance model substantially contaminate the quality of data mixture. Consequently, using twice the amount of mixture data in regurgitative training is \textit{worse} in this case, as it further accelerates the performance decline. In contrast, mixing data generated by high-performance model and GPT-3.5, both of which have good quality, results in higher performance than both constituent models. In fact, regurgitative training with the data mixture matches the performance of training with real data when the same amount of data is used, and in fact \textit{outperforms} it when twice the amount is used. The performance gain is especially evident when training the low-performance baseline model (right plot of Figure \ref{fig:mitigation_mix_hg}). Finally, we note that the data generated by high-performance model and GPT-3.5 have very similar quality (BLEU difference smaller than 0.3\%). Therefore, the encouraging results from the high-performance model / GPT-3.5 mixture also lend additional support for our mechanism analyses in Section \ref{sec:mechanism} -- errors in generated data are not the only factor affecting regurgitative training performance, and other factors (such as lexical diversity) can be at play.

\subsection{Mitigation Strategy based on AI Detection} \label{sec:mitigation_detection}
The abundance of AI-generated content online has prompted academia and industry to develop various methods to distinguish between human- and AI-generated content. For instance, GPTZero is a leading AI detector used to identify whether a document was written by LLMs such as ChatGPT. This inspires us to consider using AI detection tools to mitigate the harm of regurgitative training. In particular, instead of trying to ``catch" AI-generated content, we re-purpose AI detection tools to identify AI-generated content that closely resembles human-generated content. Then, we prioritize using AI-generated data that are indistinguishable from human-generated data (from the perspective of the AI detector) in regurgitative training. This mitigation strategy is essentially an ``imitation" approach -- regardless of why human-generated data are different from AI-generated data (error rates, lexical diversity, or other characteristics), AI-generated data that imitate human-generated data sufficiently well may be more advantageous for regurgitative training.

Starting from the fine-tuning setting, we train an AI detection classifier using randomly sampled 75,000 real translations and another 75,000 translations respectively generated by GPT-3.5, GPT-4, and LLAMA2 (not previously used in regurgitative fine-tuning). For each LLM, we construct a balanced labeled dataset with 150,000 instances, half of which are real human-generated translations and the other half are LLM-generated translations. Each data instance is structured as $input =(g_1, g_2, \ldots, g_M, [SEP], e_1, e_2, \ldots, e_N, [SEP]), label \in \{0,1\}$ where $label = 0$ marks that the English translation is generated by an LLM and 1 otherwise. Same as how we have trained the BLEU prediction model in the previous section, we retrieve token embeddings from multilingual BERT. For each LLM, we train a separate classifier on 80\% of labeled data and evaluate it on the remaining 20\%. Performance scores of various supervised techniques are listed in Appendix \ref{ap:AI_detection}. The Linear Discriminant Analysis (LDA) turns out to have highest predictive performance for both GPT-3.5 and GPT-4, and Logistic Regression has the best performance for LAMMA2.

\begin{figure}[!tbh]
 \centering
  {\includegraphics[width=\textwidth]{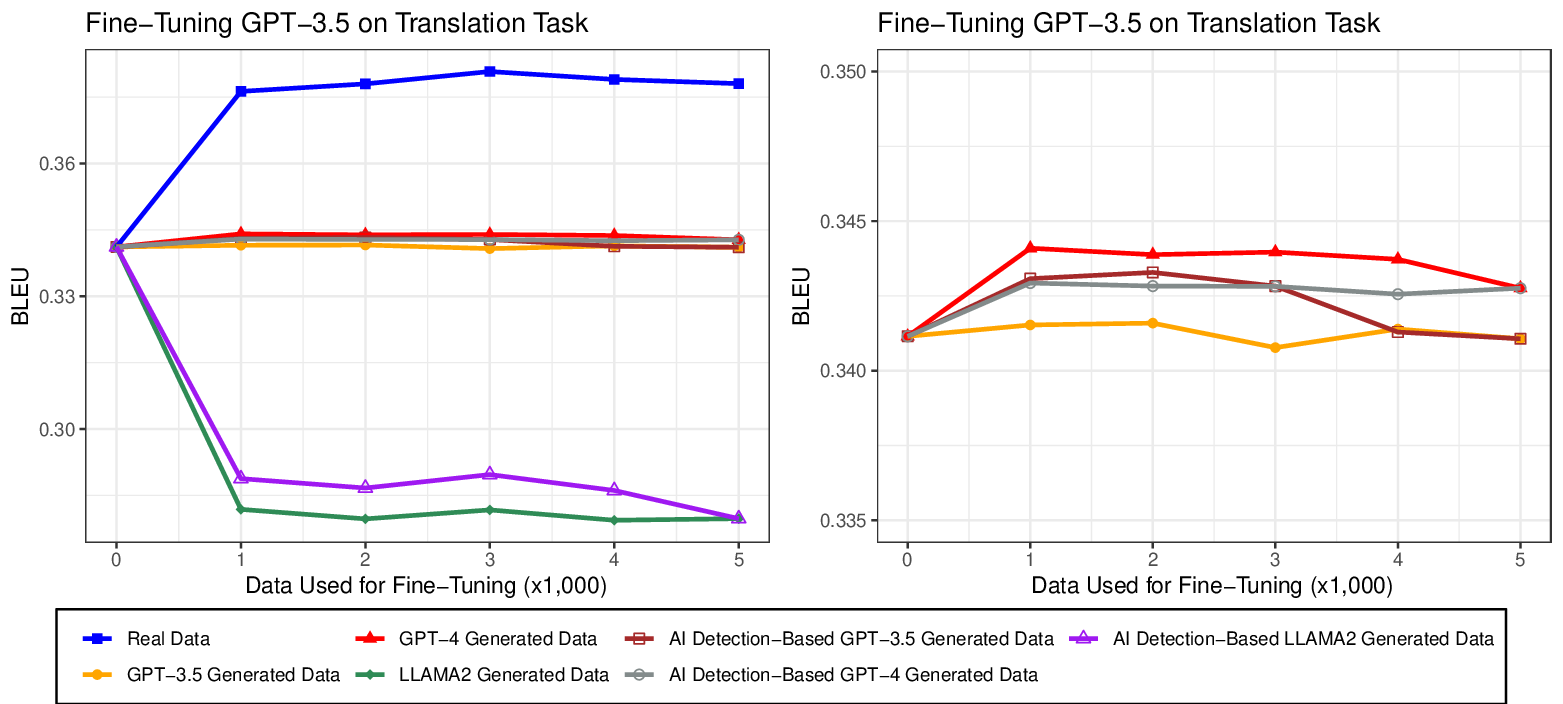}}
 \caption{AI Detection-Based Mitigation Strategy: Results on LLM Fine-Tuning (plot on the right zooms in on GPT-related results for better readability) \label{fig:finetune_detect}}
\end{figure}

Next, we apply the best-performing AI detection classifier for each of the three LLMs on the 5,000 LLM-generated translations used for regurgitative fine-tuning. Because we know the translations are generated by LLMs, if a translation receives a higher class 1 predicted probability from the AI detection classifier, then it has a greater resemblance to real translation. Therefore, we carry out regurgitative training by adding LLM-generated data in the order of their class 1 predicted probabilities, from high to low. Other experiment settings are kept the same as in Section \ref{sec:experiments_finetune}, and the results are shown in Figure \ref{fig:finetune_detect}.

We can see that regurgitative training in the order of resemblance with real data can indeed mitigate the performance loss for LLAMA2 and GPT-3.5, and the performance gain is larger for LLAMA2. This strategy, however, does not seem to be effective for GPT-4. We suspect such variations in mitigation effectiveness have to do with the capability of AI detection classifier -- indeed, AI detection is most accurate for LLAMA2-generated data and least accurate for GPT-4-generated data (see Appendix \ref{ap:AI_detection}). 

\begin{figure}[!tbh]
 \centering
  {\includegraphics[width=\textwidth]{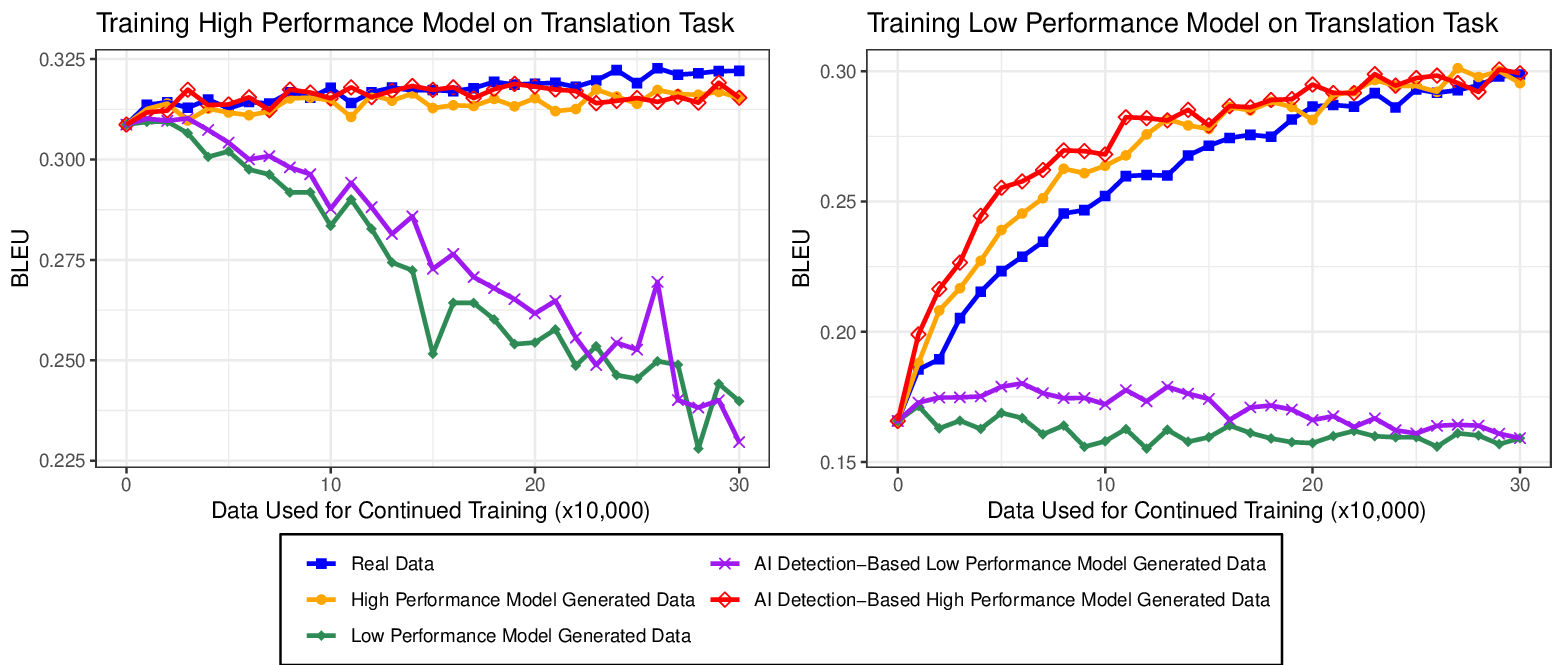}}
 \caption{AI Detection-Based Mitigation Strategy: Results on Training Transformer Models from Scratch (two plots have different $y$-axis scales for better readability) \label{fig:mitigation_llama2_detection}}
\end{figure}

Next, we apply the same mitigation strategy on transformer models trained from scratch. Performance evaluations of various AI detection classifiers are again listed in Appendix \ref{ap:AI_detection}. We choose a Logistic Regression classifier for the low-performance baseline model and a LDA classifier for the high-performance baseline model. The regurgitative training results, with data ranked by class 1 predicted probabilities, are shown in Figure \ref{fig:mitigation_llama2_detection}.

The results confirm the effectiveness of AI detection-based mitigation strategy for regurgitative training with transformers trained from scratch. Notably, different from the finding in quality-based mitigation, we observe performance gain for the high-performance baseline model as well (left plot). In fact, regurgitative training with ranked data keeps up with, and even slightly outperforms, training with real data for more than 20 batches. 

These encouraging results highlight the promising utility of AI detection outside of its conventional use case. Besides identifying AI-generated content, a capable AI detector can also be re-purposed to guide more meaningful regurgitative training. Meanwhile, we acknowledge that this mitigation strategy clearly does not have unlimited capacity. If synthetic data are generated by a less capable model (specifically, LLAMA2 in the fine-tuning setting and the low-performance model in the training-from-scratch setting), even the AI detection-enhanced regurgitative training (marked by purple lines in Figures \ref{fig:finetune_detect}-\ref{fig:mitigation_llama2_detection}) cannot catch up with the performance of training with real data.

\section{Discussions} \label{sec:discussion}
In 1950, Alan Turing envisioned the ``Imitation Game" (later termed the ``Turing test") as a test of intelligence, where a machine is treated as exhibiting intelligence if a questioner cannot reliably differentiate conversations generated by the machine or by a human being \citep{turing1950}. Now, popular LLMs on the market possess astounding capabilities to generate coherent texts and mirror human thoughts, leading some to argue that Turing test is no longer appropriate or sufficient to assess artificial intelligence \citep{sejnowski2023large,biever2023chatgpt}. If LLMs can already generate human-like content, a natural question to ask is whether they can effectively generate new data to keep training themselves.

Our analyses in this paper give a negative answer to this question. Training a new LLM using data generated (at least partially) by itself or other LLMs, a process we refer to as regurgitative training, generally results in lower performance than training with real data. While performance loss of regurgitative training has been documented in \cite{shumailov2023curse} with early versions of generative models (non-transformer-based models or pre-trained models before GPT-3.5), our work provides more comprehensive evidence by both fine-tuning commercial LLMs (including GPT-3.5, GPT-4, and LLAMA2) and training small-scale transformer models from scratch. Our explorations also reveal more nuanced performance effects of regurgitative training. Under both fine-tuning and training-from-scratch settings, regurgitative training with data generated by a competent model may still improve performance to a small extent (compared to the baseline performance without regurgitative training), but such performance improvement typically comes in a much lower speed / magnitude than training with real data. In contrast, regurgitative training with data generated by a poor model hurts performance (and of course also underperforms training with real data). These effects manifest even when only a small proportion of training data are synthetic. Even in the rare case where regurgitative training outperforms training with real data (i.e., training low-performance model with high-performance model generated data, Figure \ref{fig:translation_res}), such advantage disappears after a large amount of synthetic data is used. 

To make sense of the overall negative performance impact of regurgitative training, we compare the textual data generated by LLMs vs. humans. In the context of machine translation, we find supporting evidence that LLM-generated data not only contain more translation errors but also lower lexical diversity, both of which may contribute to the performance disadvantages. These findings align with multiple recent research that documents a ``diversity shortage" of LLM-generated content \citep[e.g.,][]{padmakumar2023does,doshi2023generative,anderson2024homogenization,zhou2024generative}, and associate it with the performance loss of regurgitative training.

These explorations of underlying mechanisms also produce potential strategies to mitigate performance loss of regurgitative training. In total, we propose and test three different strategies, respectively designed to address the issues of data quality / error, lack of lexical diversity, and both. The quality-based mitigation strategy prioritizes the use of high-quality data for regurgitative training, where ``quality" can be gauged either by prediction confidence or via a supervised learning approach. The data mixture strategy seeks to enhance lexical diversity by mixing together data generated from different LLMs. The AI detection-based strategy re-purposes an AI detection classifier to identify LLM-generated data that resemble real data, then prioritize their use in regurgitative training. While all three strategies can reduce performance loss to some extent, their relative effectiveness demonstrates some interesting nuances. First, they tend to be more effective on transformer models trained from scratch than fine-tuned LLMs. Under the fine-tuning setting, none of the migration strategies can bridge the performance gap between regurgitative training and training with real data; in contrast, when applied on the low-performance baseline model, the data mixture and AI detection based strategies can \textit{outperform} training with real data on models trained from scratch. Second, the data mixture strategy does not perform well if constituent LLMs differ both in terms of quality and lexical diversity -- the drop in data quality offsets the benefits of increased diversity. Instead, this strategy performs much better if constituent LLMs have comparable quality but still contribute diversity benefits (e.g., using competent models trained on different data or architectures). Finally, success of the AI detection strategy hinges on the ability to differentiate LLM- vs. human-generated data. Greater capability in AI detection generally results in better regurgitative training performance.

Several implications for both researchers and practitioners working with LLMs are worth noting. First, we urge caution when utilizing synthetic LLM-generated data when training or fine-tuning LLMs. Despite the multitude of amazing capabilities of LLMs, at their current stage, regurgitative training cannot create sustained performance improvement and often deteriorates the model's performance. Therefore, datasets that are organically generated and carefully curated (such as the Europarl corpus for machine translation and SQuAD corpus for Q\&A) remain part of the core assets of LLM development. Second, the prevalent use of LLMs implies that LLM-generated data would likely take up a non-trivial proportion of online content in the near future, and some degree of regurgitative training may be unavoidable. Recognizing this trend, we advocate for more careful use of LLM-generated data. Our results suggest that ``data quality" trumps ``data quantity" in regurgitative training -- it is generally more advantageous to use data with higher prediction confidence, greater linguistic richness, and higher resemblance to real data than merely using a larger quantity of data with questionable quality. Moreover, the baseline performance of the model being regurgitatively trained also matters. All else being equal, a more capable baseline model suffers less performance loss due to regurgitative training. Therefore, it is important for researchers and businesses to first thoroughly train their baseline models before engaging in regurgitative training, which helps control the adverse impact of regurgitative training. Finally, human-generated data often reflect human needs, preferences, and value judgments (e.g., online reviews reflect user preferences about products), whereas LLM-generated data may not necessarily reflect the same degree of genuine human needs and value judgments. Consequently, regurgitative training can also impact a business' ability to transform data into value. Assessing the nature of this impact and designing mitigation strategies to manage it is also an important organizational objective.

Our work also opens up a few interesting future research directions. Capabilities and performance of modern LLMs are constantly evolving. Is the negative performance impact of regurgitative training just a transient pattern reflecting limitations of available LLMs (which may disappear as more powerful LLMs are created in the future), or is it a fundamental issue of current paradigm of generative AI? Existing work such as \cite{shumailov2023curse} and \cite{gerstgrasser2024model} attempt to answer this question by resorting to analyzing simplified models (e.g., one-dimensional Gaussian processes). Future work can employ more advanced theoretical frameworks to derive deeper understandings. Moreover, effective regurgitative training with synthetic data represents an emerging field of increasing importance. Our proposed mitigation strategies are only the first steps rather than final words, and we encourage future work to design more potent strategies that can be adopted in practice. We believe more productive use of synthetic data for LLM training requires both theoretical understanding of its performance upper bound, as well as practical algorithms and methods to achieve what is possible. Lastly, LLMs are used not only in tasks with established performance metrics (such as the machine translation and Q\&A tasks considered in this work), but also in open-ended tasks whose quality are very hard to gauge (e.g., creative ideation). How to conceptually think about and empirically evaluate the impact of regurgitative training in these open-ended tasks remains an interesting question.

\bibliography{SSRN.bbl}

\clearpage
\appendix

\section{Robustness Check: Regurgitative Training with A Fixed Percentage of Training Data} \label{ap:percentage}
\begin{figure}[!tbh]
 \centering
  {\includegraphics[width=\textwidth]{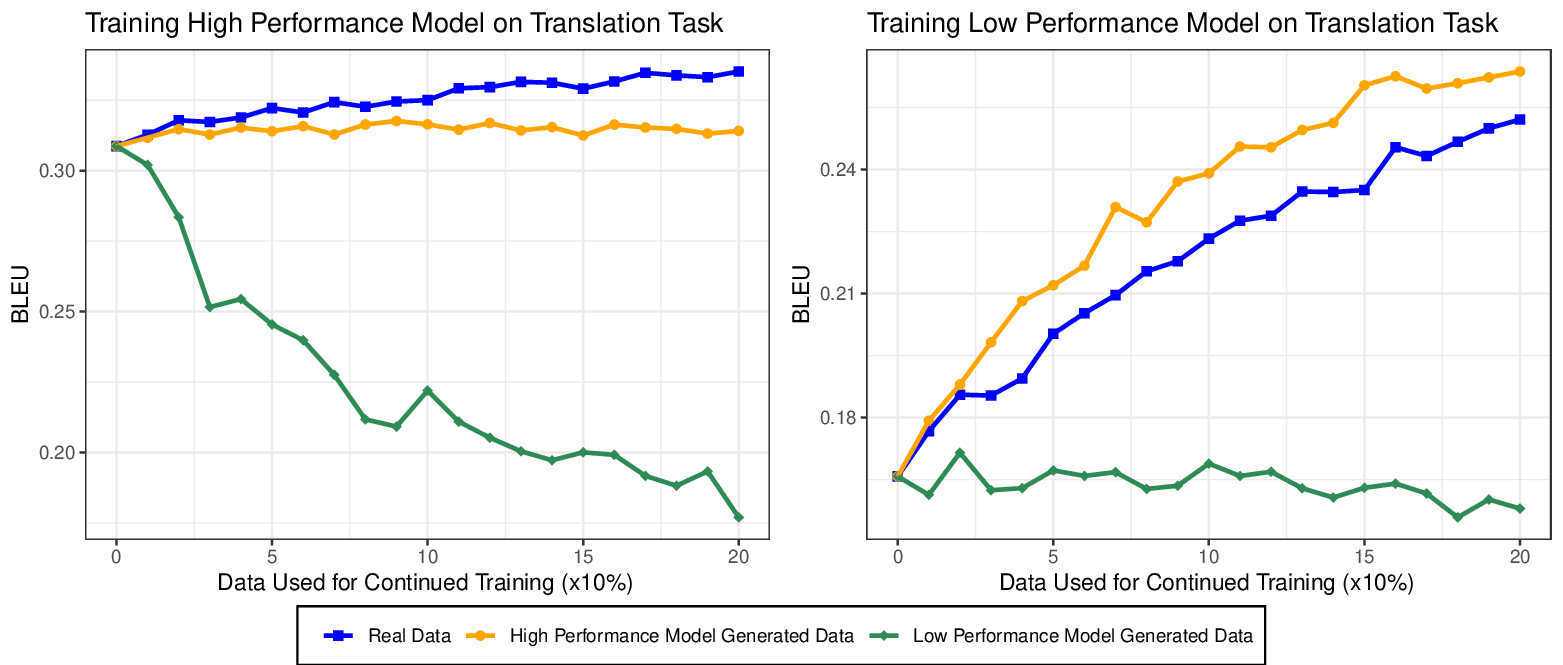}}
 \caption{Performance of Regurgitative Training Transformer Models with A Fixed Percentage of Training Data (two plots have different $y$-axis scales for better readability) \label{fig:translation_res_percentage}}
\end{figure}

\begin{figure}[!tbh]
 \centering
  {\includegraphics[width=\textwidth]{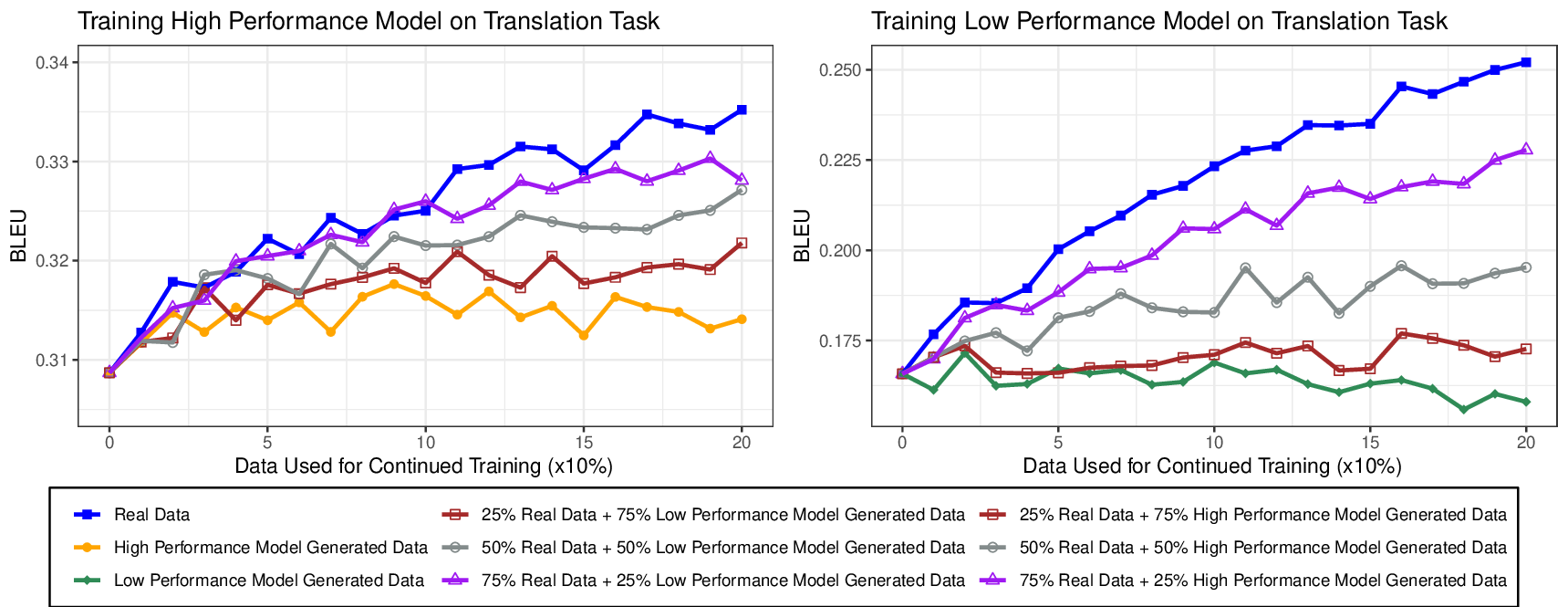}}
 \caption{Performance of Regurgitative Training Transformer Models with Different Proportions of Real Data and A Fixed Percentage of Training Data (two plots have different $y$-axis scales for better readability) \label{fig:translation_res1_percentage}}
\end{figure}

Figure \ref{fig:translation_res_percentage} shows the results of regurgitative training on both low- and high-performance baseline models with model-generated data that amount to 10\% the size of their respective training data. Figure \ref{fig:translation_res1_percentage} shows the results when a certain percentage of real data is mixed in.

\section{Identifying Low-/High-Performance Models in Q\&A Task} \label{ap:QA_real}
\begin{figure}[!tbh]
 \centering
 {\includegraphics[width=0.9\textwidth]{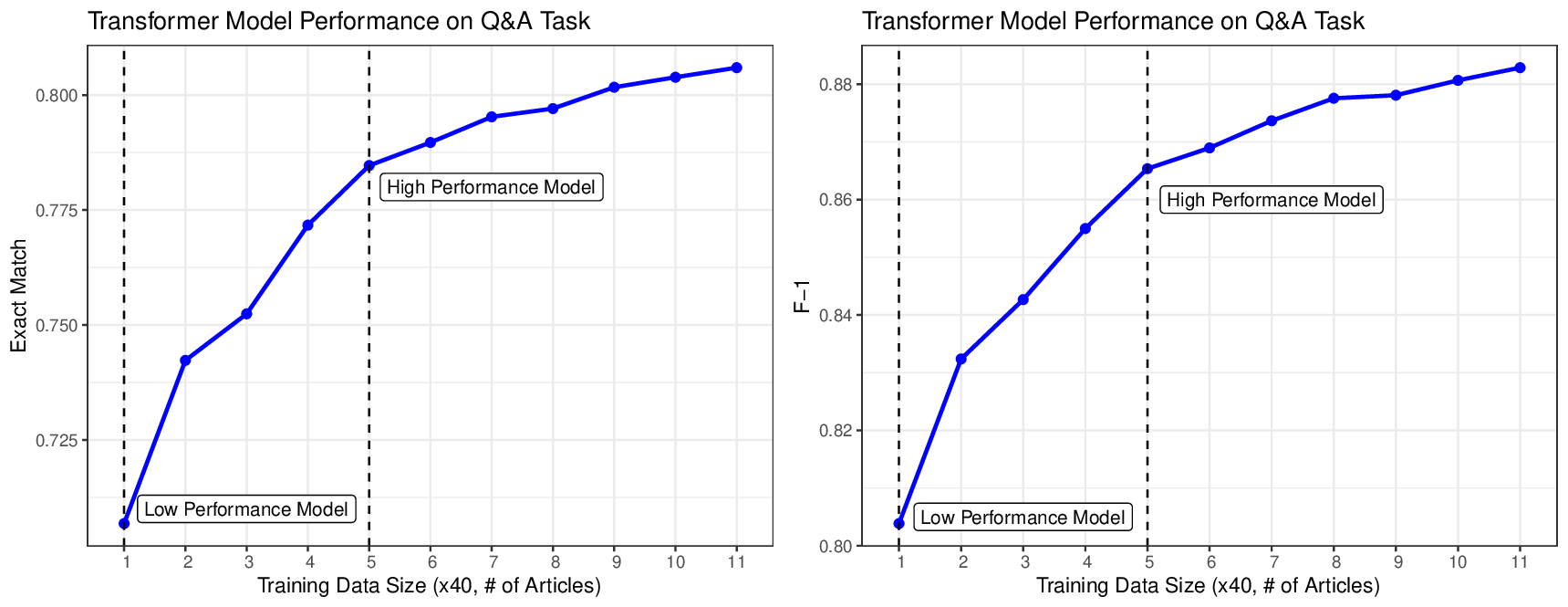}}
 \caption{Performance of Transformer Models on Q\&A Task with Varying Training Data Sizes \label{fig:qa_real}}
\end{figure}

We partition the articles in the training set of SQuAD randomly into 11 batches of data (each containing 40 articles). We incrementally add training data, one batch at a time, and evaluate the model performance on the testing data. Figure \ref{fig:qa_real} shows the performance in terms of exact match and average F-1 score. Based on these results, we choose the model trained with one batch of data as the low-performance baseline, and the model trained with five batches of data as the high-performance baseline.

\section{BLEU Prediction Performance in Quality-Based Mitigation Strategy} \label{ap:quality_bleu_prediction}
The following Tables \ref{gptperformance}-\ref{llama2performance} summarize the BLEU prediction performance of various supervised techniques for translations generated by GPT-3.5, GPT-4, and LLAMA2, respectively.

\begin{table}[!htbp]
    \centering 
    \caption{Performance of BLEU Prediction Models with GPT-3.5 Data}\label{gptperformance}    
	\begin{tabular}{cccc}
		\toprule
		 Model  & MSE & RMSE& MAE\\
		\midrule 
		\textbf{Bayesian Ridge} & \textbf{0.0415} & \textbf{0.2038}& \textbf{0.1602} \\
		Ridge Regression& 0.0416 & 0.2038&  0.1601 \\
        Linear Regression & 0.0416 & 0.2039 & 0.1602\\
        Light Gradient Boosting Machine & 0.0417  &0.2043& 0.1608  \\
         Orthogonal Matching Pursuit   & 0.0428 & 0.2069&0.1632 \\ 
          Extra Trees Regressor & 0.0421 & 0.2051  & 0.1613\\
  Gradient Boosting Regressor & 0.0427 & 0.2067 & 0.1636 \\
 Extreme Gradient Boosting & 0.0430 & 0.2074  & 0.1617 \\
		\bottomrule 
	\end{tabular}
\end{table}

\begin{table}[!htbp]
    \centering 
    \caption{Performance of BLEU Prediction Models with GPT-4 Data}\label{GPT4performance}
	\begin{tabular}{cccc}
		\toprule
		 Model & MSE & RMSE & MAE\\
		\midrule 
		\textbf{Bayesian Ridge} & \textbf{0.0429} & \textbf{0.2072}& \textbf{ 0.1627}\\
		Ridge Regression&  0.0430 & 0.2072&  0.1627 \\
        Linear Regression & 0.0430  &0.2073&  0.1627 \\
        Light Gradient Boosting Machine  & 0.0430 & 0.2073&  0.1631\\
         Orthogonal Matching Pursuit   & 0.0442 & 0.2103 &0.1657\\ 
          Extra Trees Regressor & 0.0434 & 0.2083&  0.1636 \\
  Gradient Boosting Regressor& 0.0440 & 0.2098  &   0.1659  \\
 Extreme Gradient Boosting & 0.0445 & 0.2109  & 0.1643\\
		\bottomrule 
	\end{tabular}
\end{table}

\begin{table}[!htbp]
    \centering 
    \caption{Performance of BLEU Prediction Models with LLAMA2 Data}\label{llama2performance}
	\begin{tabular}{cccc}
		\toprule
		 Model & MSE & RMSE & MAE\\
		\midrule 
		\textbf{Bayesian Ridge} &\textbf{ 0.0301} & \textbf{0.1735}&  \textbf{0.1350}\\
		Ridge Regression&  0.0301 & 0.1735&  0.1350 \\
        Linear Regression & 0.0301  &0.1736&  0.1349 \\
        Light Gradient Boosting Machine   &0.0306&  0.1749& 0.1374\\
         Orthogonal Matching Pursuit &0.0311 & 0.1764 &0.1381  \\ 
          Extra Trees Regressor & 0.0314&  0.1771  &  0.1397\\
  Gradient Boosting Regressor&  0.0314 & 0.1772 &  0.1400 \\
 Extreme Gradient Boosting   &0.0314  &0.1772& 0.1371  \\
		\bottomrule 
	\end{tabular}
\end{table}

\section{Quality-Based Mitigation Strategy on Q\&A Task: Ranking by Average Entropy} \label{ap:qa_avg}
In the following Figure \ref{fig:sort_QA_avg}, we report the evaluation results of quality-based mitigation strategy on the Q\&A Task, where we prioritize articles based on the average entropy value of their constituent answers (rather than the lowest entropy value).

\begin{figure}[!tbh]
 \centering
  {\includegraphics[width=0.9\textwidth]{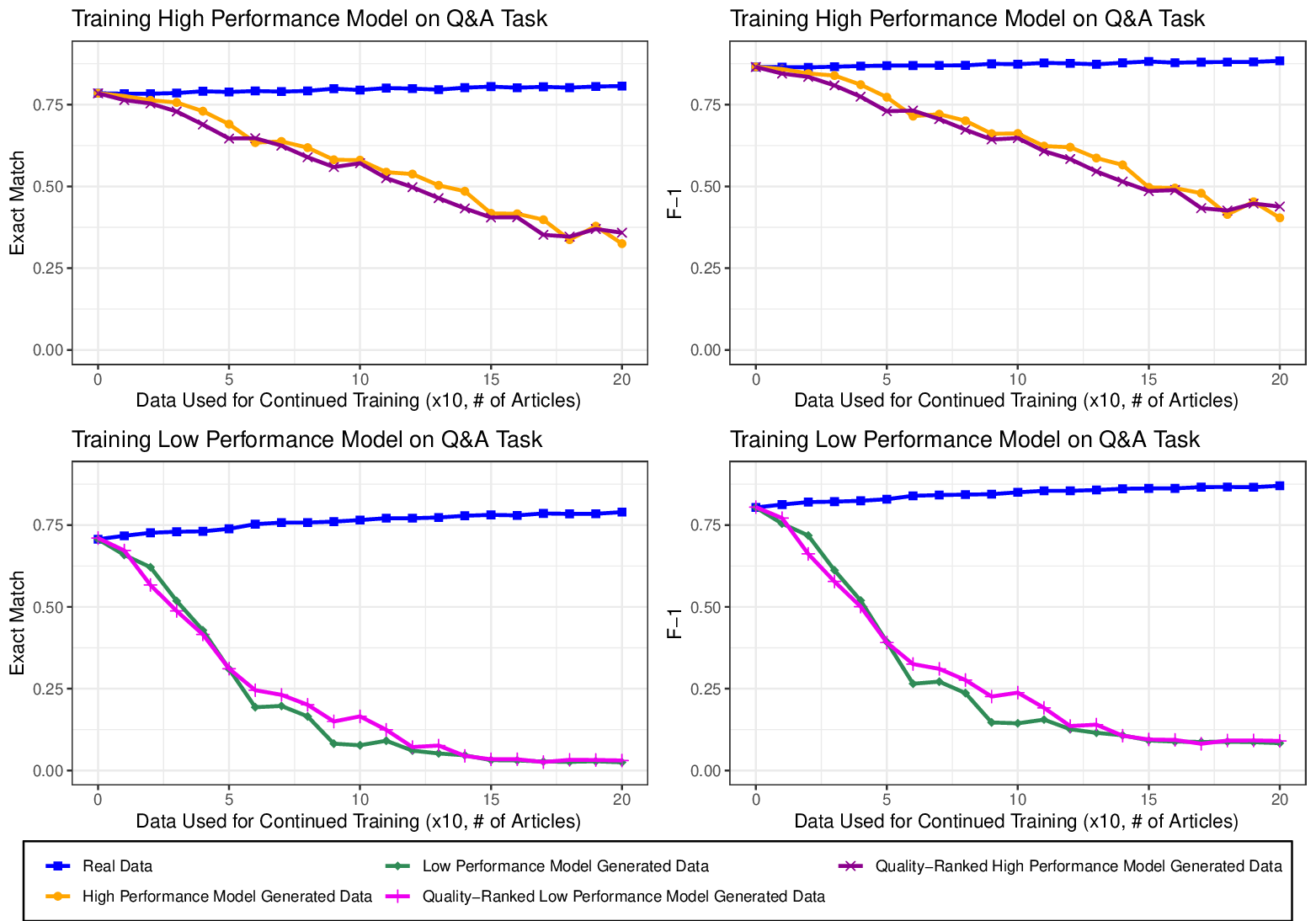}}
 \caption{Quality-Based Mitigation Strategy: Results on Q\&A Task (Ranking by Average Entropy) \label{fig:sort_QA_avg}}
\end{figure}

\section{Performance Evaluation of AI Detection Classifiers} \label{ap:AI_detection}
The following Tables \ref{gptperformance1}-\ref{llama2performance1} summarize the classification performance of AI detection models for LLM-generated translations. Tables \ref{lqperformance1}-\ref{hqperformance1} summarize the classification performance of AI detection models for translations generated by low-/high-performance transformer models.

\begin{table}[!tbh]
    \centering 
\caption{Performance of AI Detection Classifiers on GPT-3.5 Data}\label{gptperformance1}
	\begin{tabular}{ccccc}
		\toprule
		 Model & Accuracy& AUC & Recall&Precision\\
		\midrule 
  Logistic Regression & 0.6794 & 0.7468  &0.6774&  0.6796 \\
      \textbf{Linear Discriminant Analysis } &  \textbf{ 0.6800} & \textbf{0.7471} & \textbf{0.6777}&  \textbf{0.6803 } \\
     Extreme Gradient Boosting  &  0.6531 & 0.7183&  0.6495&  0.6538\\
 Light Gradient Boosting Machine  &   0.6564 & 0.7206&  0.6465 & 0.6590\\
   Random Forest   &    0.6328  &  0.6897  &  0.6197 &   0.6359\\
    Ada Boost &   0.6169 &   0.6665  &  0.6214  &  0.6153\\
    K Nearest Neighbors   &   0.6010  &  0.6433 &   0.6248  &  0.5959\\
    Naive Bayes   &   0.5879  &  0.6268  &  0.6188  &  0.5823\\
		\bottomrule 
	\end{tabular}
\end{table}

\begin{table}[!tbh]
    \centering 
    \caption{Performance of AI Detection Classifiers on GPT-4 Data}\label{gpt4performance1}
	\begin{tabular}{ccccc}
		\toprule
		 Model & Accuracy& AUC & Recall&Precision\\
		\midrule 
  Logistic Regression & 0.6778 & 0.7429 & 0.6743&  0.6791 \\
     \textbf{ Linear Discriminant Analysis}  &  \textbf{ 0.6785 } &\textbf{ 0.7446}  &\textbf{ 0.6751} &\textbf{  0.6798}\\
     Extreme Gradient Boosting  &   0.6516 & 0.7158&  0.6460 & 0.6533\\
 Light Gradient Boosting Machine  &  0.6535 & 0.7176 & 0.6419  &0.6572\\
 Random Forest &  0.6301 & 0.6851 & 0.6107 & 0.6353\\
 Ada Boost &   0.6152 & 0.6618 & 0.6094&  0.6166\\
K Nearest Neighbors &   0.5983 & 0.6380 & 0.6289  &0.5926\\
Naive Bayes  & 0.5870&  0.6250 & 0.6050 & 0.5840\\
		\bottomrule 
	\end{tabular}
\end{table}

\begin{table}[!tbh]
    \centering 
    \caption{Performance of AI Detection Classifiers on LLAMA2 Data}\label{llama2performance1}
	\begin{tabular}{ccccc}
		\toprule
		 Model & Accuracy& AUC & Recall&Precision\\
		\midrule 
  \textbf{Logistic Regression} & \textbf{0.7316} & \textbf{0.8095}  & \textbf{0.7394} & \textbf{0.7276} \\
      Linear Discriminant Analysis  &  0.7303  &0.8087&  0.7402 & 0.7254\\
     Extreme Gradient Boosting  &  0.6987  &0.7750  &0.7068&  0.6951\\
 Light Gradient Boosting Machine  &  0.6925 & 0.7659 & 0.7103 & 0.6854\\
 Random Forest & 0.6628&  0.7263 & 0.6810  &0.6566\\
 Ada Boost &   0.6469&  0.7034 & 0.6605  &0.6425\\
K Nearest Neighbors &    0.6347 & 0.6837 & 0.6507 & 0.6301\\
Naive Bayes   & 0.6104&  0.6550&  0.6631&  0.5994\\
		\bottomrule 
	\end{tabular}
\end{table}

\begin{table}[!tbh]
    \centering 
    \caption{Performance of AI Detection Classifiers on Low-Performance Transformer Data}\label{lqperformance1}
	\begin{tabular}{ccccc}
		\toprule
		 Model & Accuracy& AUC & Recall&Precision\\
		\midrule 
 \textbf{Logistic Regression}& \textbf{0.8384}& \textbf{ 0.9209}& \textbf{0.8606} &\textbf{ 0.8237}  \\
      Linear Discriminant Analysis & 0.8343  &0.9172&  0.8726 & 0.8103\\
     Extreme Gradient Boosting  &  0.7952  &0.8853 & 0.8212 & 0.7803\\
 Light Gradient Boosting Machine  &   0.7824 & 0.8707 & 0.8255 & 0.7597\\
 Random Forest &   0.7476 & 0.8306 & 0.7754  &0.7343\\
 Ada Boost &  0.7260 & 0.8064 & 0.7336  &0.7221\\
K Nearest Neighbors &   0.6906&  0.7607 & 0.6598  &0.7025\\
Naive Bayes & 0.6594&  0.7189 & 0.6778&  0.6533\\
		\bottomrule 
	\end{tabular}
\end{table}

\begin{table}[!tbh]
    \centering 
    \caption{Performance of AI Detection Classifiers on High-Performance Transformer Data}\label{hqperformance1}
	\begin{tabular}{ccccc}
		\toprule
		 Model & Accuracy& AUC & Recall&Precision\\
		\midrule 
  Logistic Regression &  0.6820 & 0.7528 & 0.7036  & 0.6740\\
  \textbf{   Linear Discriminant Analysis } & \textbf{ 0.6825} & \textbf{0.7526}  &\textbf{0.7093} & \textbf{0.6727}\\
     Extreme Gradient Boosting  &   0.6512 &   0.7159  &  0.6644  &  0.6469\\
 Light Gradient Boosting Machine  &     0.6557  &  0.7200  &  0.6759  &  0.6491\\
 Random Forest &   0.6287  &  0.6848  &  0.6268  &  0.6287\\
 Ada Boost &  0.6227  &  0.6736  &  0.6359 &   0.6190\\
K Nearest Neighbors &   0.5923   & 0.6265 &   0.5972  &  0.5909\\
Naive Bayes & 0.5919  &  0.6303  &  0.6389  &  0.5835\\
		\bottomrule 
	\end{tabular}
\end{table}

\end{document}